\documentclass{article}

\PassOptionsToPackage{numbers, compress}{natbib}

 \usepackage[final]{neurips_2023}

\usepackage[utf8]{inputenc} 
\usepackage[T1]{fontenc}    
\usepackage{hyperref}       
\usepackage{url}            
\usepackage{booktabs}       
\usepackage{amsfonts}       
\usepackage{nicefrac}       
\usepackage{microtype}      
\usepackage{xcolor}         
\usepackage{multirow}
\usepackage{multicol}
\usepackage{graphicx}
\usepackage{amsmath}
\usepackage{amsthm}
\usepackage{colortbl}  
\usepackage{xcolor}
\usepackage{bm}
\usepackage{enumitem}
\usepackage{wrapfig}
\usepackage{amssymb}
\usepackage{subfigure}
\usepackage{calc}
\usepackage{algorithm}
\usepackage{algorithmic}
\PassOptionsToPackage{numbers, compress}{natbib}

\title{Diffusion Model is an Effective Planner and Data Synthesizer for Multi-Task Reinforcement Learning}

\author{%
  \textbf{Haoran He$^{\,1,2}$\thanks{The work was conducted during the internship of Haoran He at Shanghai Artificial Intelligence Laboratory.}$\quad$
  Chenjia Bai$^2$\thanks{Corresponding authors: Chenjia Bai \href{baichenjia@pjlab.org.cn}{(baichenjia@pjlab.org.cn)}, Xuelong Li}$\quad$
  Kang Xu$^{2,3}\quad$
  Zhuoran Yang$^4\quad$
  Weinan Zhang$^1$}\\\\
  \textbf{Dong Wang$^2\quad$
  Bin Zhao$^{2,5}\quad$
  Xuelong Li$^{2,5\dag}$}
  \\\\
$^1$Shanghai Jiao Tong University$\quad$
$^2$Shanghai Artificial Intelligence Laboratory\\
$^3$Fudan University$\quad$
$^4$Yale University$\quad$
$^5$Northwestern Polytechnical University\\
}

\begin{document}

\maketitle

\begin{abstract}
Diffusion models have demonstrated highly-expressive generative capabilities in vision and NLP. Recent studies in reinforcement learning (RL) have shown that diffusion models are also powerful in modeling complex policies or trajectories in offline datasets. However, these works have been limited to single-task settings where a generalist agent capable of addressing multi-task predicaments is absent. In this paper, we aim to investigate the effectiveness of a single diffusion model in modeling large-scale multi-task offline data, which can be challenging due to diverse and multimodal data distribution. Specifically, we propose Multi-Task Diffusion Model (\textsc{MTDiff}), a diffusion-based method that incorporates Transformer backbones and prompt learning for generative planning and data synthesis in multi-task offline settings. \textsc{MTDiff} leverages vast amounts of knowledge available in multi-task data and performs implicit knowledge sharing among tasks. For generative planning, we find \textsc{MTDiff} outperforms state-of-the-art algorithms across 50 tasks on Meta-World and 8 maps on Maze2D. For data synthesis, \textsc{MTDiff} generates high-quality data for testing tasks given a single demonstration as a prompt, which enhances the low-quality datasets for even unseen tasks.
\end{abstract}

\section{Introduction}

The high-capacity generative models trained on large, diverse datasets have demonstrated remarkable success across vision and language tasks. An impressive and even preternatural ability of these models, e.g. large language models (LLMs), is that the learned model can generalize among different tasks by simply conditioning the model on instructions or prompts \cite{imageGen, dalle2,gpt3,chowdhery2022palm,touvron2023llama,openai2023gpt4,visualGPT}. 
The success of LLMs and vision models inspires us to utilize the recent generative model to learn from large-scale offline datasets that include multiple tasks for generalized decision-making in reinforcement learning (RL). Thus far, recent attempts in offline decision-making take advantage of the generative capability of diffusion models \cite{thermodynamic,ddpm} to improve long-term planning \cite{diffuser,decisiondiffuser} or enhance the expressiveness of policies \cite{diffusionql,diffusionpolicy,behaviormodeling}. However, these works are limited to small-scale datasets and single-task settings where broad generalization and general-purpose policies are not expected. In multi-task offline RL which considers learning a single model to solve multi-task problems, the dataset often contains noisy, multimodal, and long-horizon trajectories collected by various policies across tasks and with various qualities, which makes it more challenging to learn policies with broad generalization and transferable capabilities.
Gato \cite{gato} and other generalized agents \cite{gamedt,wen2022realization} take transformer-based architecture \cite{transformer} via sequence modeling to solve multi-task problems, while they are highly dependent on the optimality of the datasets and are expensive to train due to the huge number of parameters. 

To address the above challenges, we propose a novel diffusion model to further explore its generalizability in a multi-task setting. 
We formulate the learning process from multi-task data as a denoising problem, which benefits the modeling of multimodal data. Meanwhile, we develop a relatively lightweight architecture by using a GPT backbone \cite{gpt2} to model sequential trajectories, which has less computation burden and improved sequential modeling capability than previous U-Net-based \cite{unet} diffusion models \cite{diffuser,AdaptDiffuser}.
To disambiguate tasks during training and inference, instead of providing e.g. one-hot task identifiers, we leverage demonstrations as \emph{prompt} conditioning, which exploits the few-shot abilities of agents \cite{prompt4zeroshot, wei2022finetuned,promptdt}.
\begin{figure*}[t]
    \centering
    \includegraphics[width=1.0\linewidth]{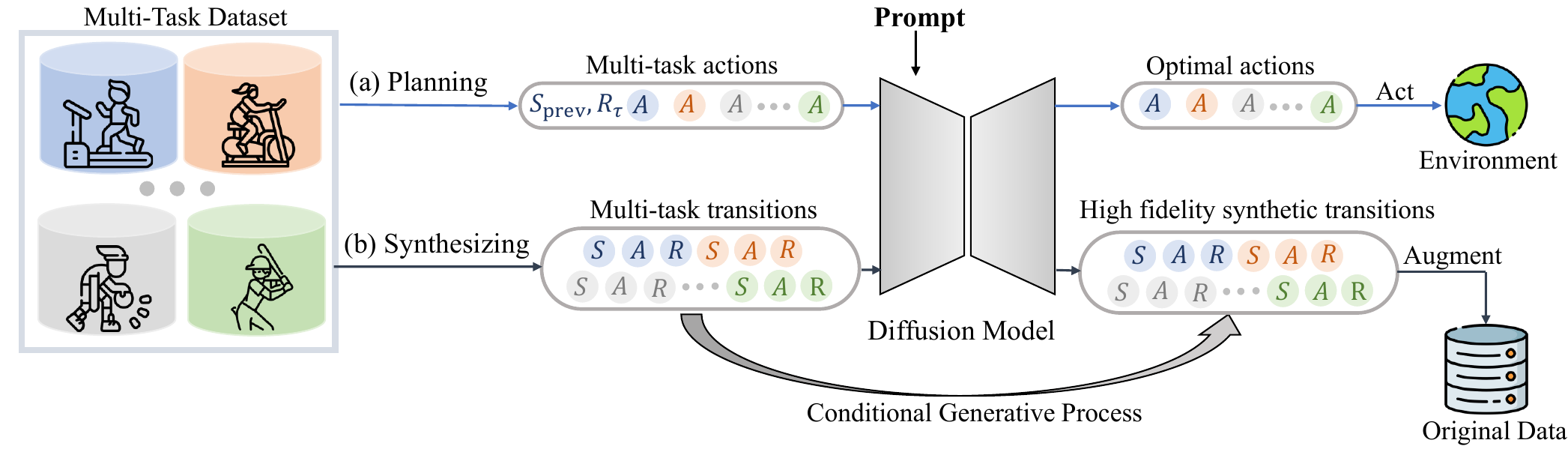}
    \caption{Overall architecture of \textsc{MTDiff}. Different colors represent different tasks. $S$, $A$ and $R$ denote the state sequence, action sequence, and reward sequence from multi-task data, respectively. $S_{\rm{prev}}$ and $R_\tau$ represent historical states and normalized return.
    }
    \label{fig:illustration_models}
\end{figure*}

We name our method the Multi-Task Diffusion Model (\textbf{\textsc{MTDiff}}). As shown in Figure \ref{fig:illustration_models}, we investigate two variants of \textsc{MTDiff} for \underline{p}lanning and data \underline{s}ynthesis to further exploit the utility of diffusion models, denoted as \textbf{\textsc{MTDiff-p}} and \textbf{\textsc{MTDiff-s}}, respectively. (a) For planning, \textsc{MTDiff-p} learns a prompt embedding to extract the task-relevant representation, and then concatenates the embedding with the trajectory's normalized return and historical states as the \emph{conditions} of the model. During training, \textsc{MTDiff-p} learns to predict the corresponding future action sequence given the conditions, and we call this process generative planning \cite{zhang2022generative}. During inference, given few-shot prompts and the desired return, \textsc{MTDiff-p} tries to denoise out the optimal action sequence starting from the current state. Surprisingly, \textsc{MTDiff-p} can adapt to unseen tasks given well-constructed prompts that contain task information. (b) By slightly changing the inputs and training strategy, we can unlock the abilities of our diffusion model for data synthesis. Our insight is that the diffusion model, which compresses informative multi-task knowledge well, is more effective and generalist than previous methods that only utilize single-task data for augmentation \cite{lu2023synthetic,sinha2022s4rl,laskin2020reinforcement}.  Specifically, \textsc{MTDiff-s} learns to estimate the joint conditional distribution of the full transitions that contain states, actions, and rewards based on the task-oriented prompt. Different from \textsc{MTDiff-p}, \textsc{MTDiff-s} learns to synthesize data from the underlying dynamic environments for each task. Thus \textsc{MTDiff-s} only needs prompt conditioning to identify tasks.
We empirically find that \textsc{MTDiff-s} synthesizes high-fidelity data for multiple tasks, including both seen and unseen ones, which can be further utilized for data augmentation to expand the offline dataset and enhance policy performance.

To summarize, \textsc{MTDiff} is a diffusion-based method that leverages the multimodal generative ability of diffusion models, the sequential modeling capability of GPT architecture, and the few-shot generalizability of prompt learning for multi-task RL. To the best of our knowledge, we are the first to achieve both effective planning and data synthesis for multi-task RL via diffusion models.
Our contributions include: (\romannumeral 1) we propose \textsc{MTDiff}, a novel GPT-based diffusion model that illustrates the supreme effectiveness in multi-task trajectory modeling for both planning and data synthesis; (\romannumeral 2) we incorporate prompt learning into the diffusion framework to learn to generalize across different tasks and even adapt to unseen tasks; (\romannumeral 3) our experiments on Meta-World and Maze2D benchmarks demonstrate that \textsc{MTDiff} is an effective planner to solve the multi-task problem, and also a powerful data synthesizer to augment offline datasets in the seen or unseen tasks.



\section{Preliminaries}

\subsection{Reinforcement Learning}

\paragraph{MDP and Multi-task MDP.} 
A Markov Decision Process (MDP) is defined by a tuple $(\mathcal{S}, \mathcal{A},\mathcal{P}, \mathcal{R}, \mathcal{\mu}, \mathcal{\gamma})$, where $\mathcal{S}$ is the state space, $\mathcal{A}$ is the action space, $\mathcal{P}:\mathcal{S}\times\mathcal{A}\to\mathcal{S}$ is the transition function, $\mathcal{R}:\mathcal{S}\times\mathcal{A}\times\mathcal{S}\to\mathbb{R}$ is the reward function for any transition, $\mathcal{\gamma}\in(0, 1]$ is a discount factor, and $\mathcal{\mu}$ is the initial state distribution. At each timestep $t$, the agent chooses an action $a_t$ by following the policy $\pi:\mathcal{S}\to\Delta_\mathcal{A}$. Then the agent obtains the next state $s_{t+1}$ and receives a scalar reward $r_t$. In single-task RL, the goal is to learn a policy $\pi^*=\arg\max_\pi\mathbb{E}_{a_t \sim \pi}\big[\sum\nolimits_{t=0}^{\infty}\gamma^t r_t\big]$ by maximizing the expected cumulative reward of the corresponding task. 

In a multi-task setting, different tasks can have different reward functions, state spaces and transition functions. We consider all tasks to share the same action space with the same embodied agent. Given a specific task $\mathcal{T} \sim p(\mathcal{T})$, a task-specified MDP can be defined as $(\mathcal{S}^{\mathcal{T}}, \mathcal{A},\mathcal{P}^{\mathcal{T}}, \mathcal{R}^{\mathcal{T}}, \mathcal{\mu}^{\mathcal{T}}, \mathcal{\gamma})$. Instead of solving a single MDP, the goal of multi-task RL is to find an optimal policy that maximizes expected return over all the tasks: $\pi^*=\arg\max_\pi\mathbb{E}_{{\mathcal{T}}\sim p(\mathcal{T})}\mathbb{E}_{a_t\sim\pi^{\mathcal T}}\big[\sum\nolimits_{t=0}^{\infty}\gamma^t r_t^{\mathcal{T}}\big]$. 

\paragraph{Multi-Task Offline Decision-Making.} In offline decision-making, the policy is learned from a static dataset of transitions $\{(s_j, a_j, s'_j, r_j)\}_{j=1}^{N}$ collected by an unknown behavior policy $\pi_\beta$ \cite{offlinerl}. In the multi-task offline RL setting, the dataset $\mathcal{D}$ is partitioned into per-task subsets as $\mathcal{D}=\cup_{i=1}^N \mathcal{D}_i$, where $\mathcal{D}_i$ consists of experiences from task $\mathcal{T}_i$. The key issue of RL in the offline setting is the distribution shift problem caused by temporal-difference (TD) learning. In our work, we extend the idea of Decision Diffuser \cite{decisiondiffuser} by considering multi-task policy learning as a conditional generative process without fitting a value function. The insight is to take advantage of the powerful distribution modeling ability of diffusion models for multi-task data, avoiding facing the risk of distribution shift.

Offline RL learns policies from a static dataset, which makes the quality and diversity of the dataset significant \cite{d4rl}. One can perform data perturbation \cite{laskin2020reinforcement} to up-sample the offline dataset. Alternatively, we synthesize new transitions $(s,a,s',r)$ by capturing the underlying MDP of a given task via diffusion models, which expands the original dataset and leads to significant policy improvement.

\subsection{Diffusion Models}
\label{sec:pre_diffusion}

We employ diffusion models to learn from multi-task data $\mathcal{D}=\cup_{i=1}^N \mathcal{D}_i$ in this paper. With $\tau$ the sampled trajectory from $\mathcal{D}$, we denote $\bm{x}_k(\tau)$ as the $k$-step denoised output of the diffusion model, and $\bm{y}(\tau)$ is the condition which represents the task attributes and the trajectory's optimality (e.g., returns).
A forward diffusion chain gradually adds noise to the data $\bm{x}_0(\tau)\sim q(\bm{x}(\tau))$ in $K$ steps with a pre-defined variance schedule $\beta_k$, which can be expressed as
\begin{equation}
\label{eq:diffusion}
    q(\bm{x}_k(\tau)|\bm{x}_{k-1}(\tau)):=\mathcal{N}(\bm{x}_k(\tau);\sqrt{1-\beta_k}\bm{x}_{k-1}(\tau), \beta_k\boldsymbol{I}).
\end{equation}
In this paper, we adopt Variance Preserving (VP) beta schedule \cite{xiao2022tackling} and define
$\beta_k = 1- \exp\big({-\beta_{min}(\frac{1}{K})-0.5(\beta_{\rm max}-\beta_{\rm min})\frac{2k-1}{K^2}}\big)$,
where $\beta_{\rm max}=10$ and $\beta_{\rm min}=0.1$ are constants. A trainable reverse diffusion chain, constructed as $p_\theta(\bm{x}_{k-1}(\tau)|\bm{x}_k(\tau),\bm{y}(\tau)):=\mathcal{N}(\bm{x}_{k-1}(\tau)|\mu_\theta(\bm{x}_k(\tau), \bm{y}(\tau), k),\Sigma_k)$, can be optimized by a simplified surrogate loss \cite{ddpm}:
\begin{equation}
    \mathcal{L}_{\rm denoise}:=\mathbb{E}_{k\sim \mathcal{U}(1, K), \bm{x}_0(\tau)\sim q,\epsilon\sim\mathcal{N(\boldsymbol{\mathrm{0}},\boldsymbol{{I}})}}[\big\|\epsilon-\epsilon_\theta(\bm{x}_k(\tau), \bm{y}(\tau), k)\big\|^2],
\end{equation}
where $\epsilon_\theta$ parameterized by a deep neural network is trained to predict the noise $\epsilon\sim\mathcal{N}(\boldsymbol{\mathrm{0}},\boldsymbol{{I}})$ added to the dataset sample $\bm{x}_0(\tau)$ to produce $\bm{x}_k(\tau)$. By setting $\alpha_k:=1-\beta_k$ and $\bar{\alpha}_k:=\prod_{s=1}^k\alpha_s$, we obtain 
\begin{equation}\nonumber
    \bm{x}_{k-1}(\tau)\!\gets\! \frac{1}{\sqrt{\alpha_k}} \left(\bm{x}_k(\tau)\!-\!\frac{\beta_k}{\sqrt{1-\bar{\alpha}_k}}\epsilon_\theta(\bm{x}_k(\tau),\bm{y}(\tau),k)\right) + \sqrt{\beta_k}\sigma,\ \sigma\!\sim\!\mathcal{N}(\boldsymbol{0},\boldsymbol{I}),\ \text{for}\ k=\{K,...,1\}.
\end{equation}

Classifier-free guidance \cite{ho2021classifierfree} aims to learn the conditional distribution $q(\bm{x}(\tau)|\bm{y}(\tau))$ without separately training a classifier. In the training stage, this method needs to learn both a conditional $\epsilon_\theta(\bm{x}_k(\tau), \bm{y}(\tau), k)$ and an unconditional $\epsilon_\theta(\bm{x}_k(\tau), \varnothing, k)$ model, where $\bm{y}(\tau)$ is dropped out. Then the perturbed noise $\epsilon_\theta(\bm{x}_k(\tau), \varnothing, k)+\alpha(\epsilon_\theta(\bm{x}_k(\tau), \bm{y}(\tau), k)-\epsilon_\theta(\bm{x}_k(\tau),\varnothing, k))$ is used to generate samples latter, where $\alpha$ can be recognized as the guidance scale.

\section{Methodology}



\subsection{Diffusion Formulation}
\label{sec:diff-ap}

To capture the multimodal distribution of the trajectories sampled from multiple MDPs, we formulate the multi-task trajectory modeling as a conditional generative problem via diffusion models: 
\begin{equation}
\label{eq:c-diffusion}
    \max\nolimits_\theta\mathbb{E}_{\tau\sim\cup_i\mathcal{D}_i}\big[\log p_{\theta}(\bm{x}_{0}(\tau)\:\big|\: \bm{y}(\tau)\big],
\end{equation}
where $\bm{x}_0(\tau)$ is the generated desired sequence and $\bm{y}(\tau)$ is the condition.
$\bm{x}_0(\tau)$ will then be used for generative planning or data synthesis through conditional reverse denoising process $p_\theta$ for specific tasks. 
Maximizing Eq.~\eqref{eq:c-diffusion} can be approximated by maximizing a variational lower bound \cite{ddpm}.
\begin{figure*}[t]
    \centering
    \includegraphics[width=1.0\linewidth]{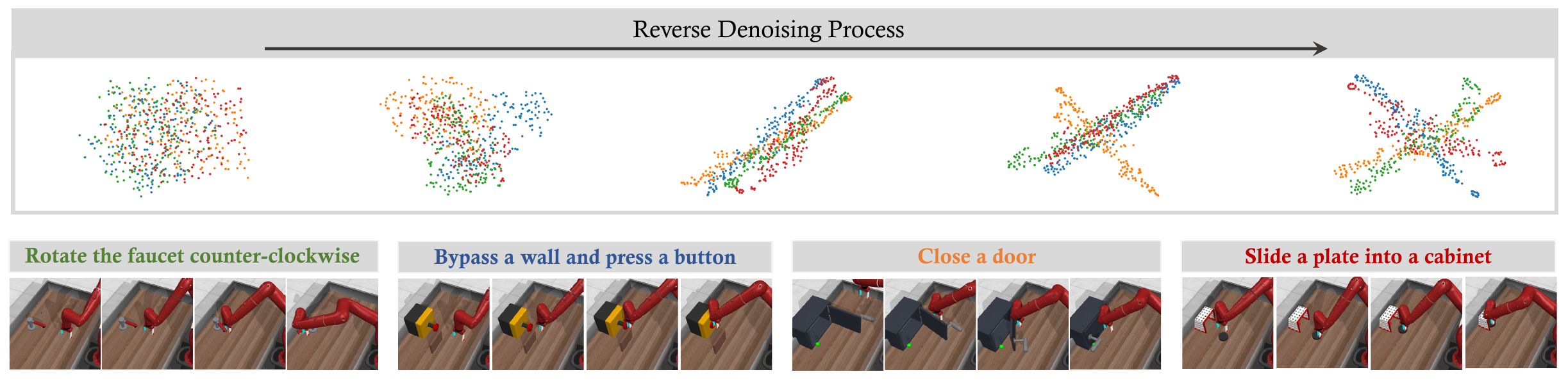}
    \caption{An example of the denoising process of \textsc{MTDiff}. We choose 4 tasks for visualization and $\bm{x}_K(\tau)$ is sampled from the Gaussian noise for each task. Since different tasks require different manipulation skills, the corresponding action sequences are dispersed in the embedding space. Our model learns such properties and generates task-specific sequences based on task-relevant prompts.
    }
    \label{fig:mt-prompt}
\end{figure*}

In terms of different inputs and outputs in generative planning and data synthesis, $\bm{x}(\tau)$ can be represented in different formats. We consider two choices to formulate $\bm{x}(\tau)$ in \textsc{MTDiff-p} and \textsc{MTDiff-s}, respectively. (\romannumeral 1)~For \textbf{\textsc{MTDiff-p}}, $\bm{x}(\tau)$ represents the action sequence for planning. We model the action sequence defined as:
\begin{equation}
\bm{x}^{\textcolor{red}{p}}_k(\tau):=(a_t, a_{t+1},...,a_{t+H-1})_k,
\end{equation}
with the context condition as 
\begin{equation}
\label{eq:y-p}
\bm{y}^{\textcolor{red}{p}}(\tau):=\big[\bm{y}'(\tau),R(\tau)\big], \quad \ \ \bm{y}'(\tau):=(Z,s_{t-L+1},...,s_{t}),
\end{equation}
where $t$, $H$, $R(\tau)$ and $L$ denote the time visited in trajectory $\tau$, the length of the input sequence $\bm{x}$, the normalized cumulative return under $\tau$ and the length of the observed state history, respectively. $Z$ is the task-relevant information as \emph{prompt}. We use $\bm{y}'(\tau)$ as an ordinary condition that is injected into the model during both training and testing, while considering $R(\tau)$ as the classifier-free guidance to obtain the optimal action sequence for a given task. (\romannumeral 2) For data synthesis in \textbf{\textsc{MTDiff-s}}, the inputs and outputs become the transition sequence that contains states, actions, and rewards, and then the outputs are utilized for data augmentation. We define the transition sequence as:
\begin{equation}
\bm{x}^{\textcolor{red}{s}}_k(\tau):=\begin{bmatrix}
s_t &s_{t+1}& \cdots & s_{t+H-1} \\
a_t &a_{t+1}&\cdots&a_{t+H-1} \\
r_t &r_{t+1}& \cdots & r_{t+H-1}
\end{bmatrix},
\end{equation}
with the condition:
\begin{equation}
\bm{y}^{\textcolor{red}{s}}(\tau):=[Z],
\end{equation}
where $\bm{y}^{\textcolor{red}{s}}(\tau) $ takes the same conditional approach as $y'(\tau)$. Figure~\ref{fig:mt-prompt} illustrates the reverse denoising process of \textsc{MTDiff-p} learned on multi-task datasets collected in Meta-World \cite{metaworld}. The result demonstrates that our diffusion model successfully distinguishes different tasks and finally generates the desired $\bm{x}_0(\tau)$. We illustrate the data distribution of $\bm{x}_0(\tau)$ in a 2D space with dimensional reduction via T-SNE \cite{tsne}, as well as the rendered states after executing the action sequence. The result shows that, with different task-specific prompts as conditions, the generated planning sequence for a specific task will be separate from sequences of other tasks, which verifies that \textsc{MTDiff} can learn the distribution of multimodal trajectories based on $\bm{y}(\tau)$. 

\subsection{Prompt, Training and Sampling}
\label{sec:condition}

In multi-task RL and LLM-driven decision-making, existing works use one-hot task identifiers \cite{sun2022paco,PCG_grad} or language descriptions \cite{saycan,brohan2022rt1} as conditions in multi-task training. Nevertheless, we argue that the one-hot encoding for each task \cite{mt-opt,bridge_data} suffices for learning a repertoire of training tasks while 
cannot generalize to novel tasks since it does not leverage semantic similarity between tasks.
In addition, the language descriptions \cite{ALFRED20,saycan,brohan2022rt1,calvin,LCR_IL} of tasks require large amounts of human labor to annotate and encounter challenges related to ambiguity \cite{yu2023using}. In \textsc{MTDiff}, we use expert demonstrations consisting of a few trajectory segments to construct more expressive prompts in multi-task settings. The incorporation of prompt learning improves the model's ability for generalization and extracting task-relevant information to facilitate both generative planning and data synthesis. We remark that a similar method has also been used in PromptDT \citep{promptdt}. Nonetheless, how such a prompt contributes within a diffusion-based framework remains to be investigated.

Specifically, we formulate the task-specific label $Z$ as trajectory prompts that contain states and actions:
\begin{equation}
    Z:=\begin{bmatrix}
    s^*_{i}&s^*_{i+1}& \cdots& s^*_{i+J-
    1}\\
    a^*_{i}&a^*_{i+1}&\cdots&a^*_{i+J-1}
    \end{bmatrix},
\end{equation}
where each element with star-script is associated with a trajectory prompt, and $J$ is the number of environment steps for identifying tasks. With the prompts as conditions, \textsc{MTDiff} can specify the task by implicitly capturing the transition model and the reward function stored in the prompts for better generalization
to unseen tasks without additional parameter-tuning.

In terms of decision-making in \textsc{MTDiff-p}, we aim to devise the optimal behaviors that maximize return. Our approach is to utilize the diffusion model for action planning via classifier-free guidance \cite{ho2021classifierfree}. Formally, an optimal action sequence $\bm{x}^{\textcolor{red}{p}}_0(\tau)$ is sampled by starting with Gaussian noise $\bm{x}_K(\tau)$ and refining
$\bm{x}^{\textcolor{red}{p}}_k(\tau)$ into $\bm{x}^{\textcolor{red}{p}}_{k-1}(\tau)$ at each intermediate timestep with the perturbed noise:
\begin{equation}
    \epsilon_\theta\big(\bm{x}^{\textcolor{red}{p}}_k(\tau),\bm{y}'(\tau), \varnothing, k\big)+\alpha\big(\epsilon_\theta(\bm{x}^{\textcolor{red}{p}}_k(\tau), \bm{y}'(\tau), R(\tau), k)-\epsilon_\theta(\bm{x}^{\textcolor{red}{p}}_k(\tau), \bm{y}'(\tau), \varnothing, k)\big),
\end{equation}
where $\bm{y}'(\tau)$ is defined in Eq.~\eqref{eq:y-p}. $R(\tau)$ is the normalized return of $\tau$, and $\alpha$ is a hyper-parameter that seeks to augment and extract the best portions of trajectories in the dataset with high return. During training, we follow DDPM \cite{ddpm} as well as classifier-free guidance \cite{ho2021classifierfree} to train the reverse diffusion process $p_\theta$, parameterized through the noise model $\epsilon_\theta$, with the following loss:
\begin{equation}
    \mathcal{L}^{\textcolor{red}{p}}(\theta):=\mathbb{E}_{k\sim \mathcal{U}(1, K), \bm{x}_0(\tau)\sim q,\epsilon\sim\mathcal{N(\boldsymbol{\mathrm{0}},\boldsymbol{{I}})},\beta\sim \rm{Bern}(p)}\big[\big\|\epsilon-\epsilon_\theta\big(\bm{x}_k^{\textcolor{red}{p}}(\tau),\bm{y}'(\tau),(1-\beta)R(\tau)+\beta\varnothing, k\big)\big\|^2\big].
\end{equation}
Note that with probability $p$ sampled from a Bernoulli distribution, we ignore the conditioning return $R(\tau)$. During inference, we adopt the \emph{low-temperature sampling} technique \cite{decisiondiffuser} to produce high-likelihood sequences. We sample $\bm{x}^{\textcolor{red}{p}}_{k-1}(\tau)\sim\mathcal{N}(\mu_\theta(\bm{x}^{\textcolor{red}{p}}_{k-1},\bm{y}'(\tau), R_{\rm max}(\tau), k-1),\beta\Sigma_{k-1})$, where the variance is reduced by $\beta\in[0,1)$ for generating action sequences with higher optimality.

For \textsc{MTDiff-s}, since the model aims to synthesize diverse trajectories for data augmentation, which does not need to take any guidance like $R(\tau)$, we have the following loss:
\begin{equation}
    \mathcal{L}^{\textcolor{red}{s}}(\theta):=\mathbb{E}_{k\sim \mathcal{U}(1, K), \bm{x}_0(\tau)\sim q,\epsilon\sim\mathcal{N(\boldsymbol{\mathrm{0}},\boldsymbol{{I}})}}\big[\big\|\epsilon-\epsilon_\theta(\bm{x}_k^{\textcolor{red}{s}}(\tau),\bm{y}^{\textcolor{red}{s}}(\tau), k)\big\|^2\big].
\end{equation}
We sample $\bm{x}^{\textcolor{red}{s}}_{k-1}(\tau)\sim\mathcal{N}(\mu_\theta(\bm{x}^{\textcolor{red}{s}}_{k-1},\bm{y}^{\textcolor{red}{s}}(\tau), k-1),\Sigma_{k-1})$. The evaluation process is given in Fig.~\ref{fig:arc}.

\subsection{Architecture Design}
\label{sec:train_arc}


Notably, the emergence of Transformer \cite{transformer} and its applications on generative modeling \cite{Peebles2022DiT,bao2023transformer, gato,brohan2022rt1} provides a promising solution to capture interactions between modalities of different tasks. Naturally, instead of U-Net \cite{unet} which is commonly used in previous single-task diffusion RL works \cite{diffuser,decisiondiffuser,AdaptDiffuser}, we parameterize $\epsilon_\theta$ with a novel transformer architecture. We adopt GPT2 \cite{gpt2} architecture for implementation, which excels in sequential modeling and offers a favorable balance between performance and computational efficiency.
Our key insight is to train the diffusion model in a unified manner to model multi-task data, treating different inputs as tokens in a unified architecture, which is expected to enhance the efficiency of diverse information exploitation. 

\begin{figure*}[t]
    \centering
    \includegraphics[width=0.9\linewidth]{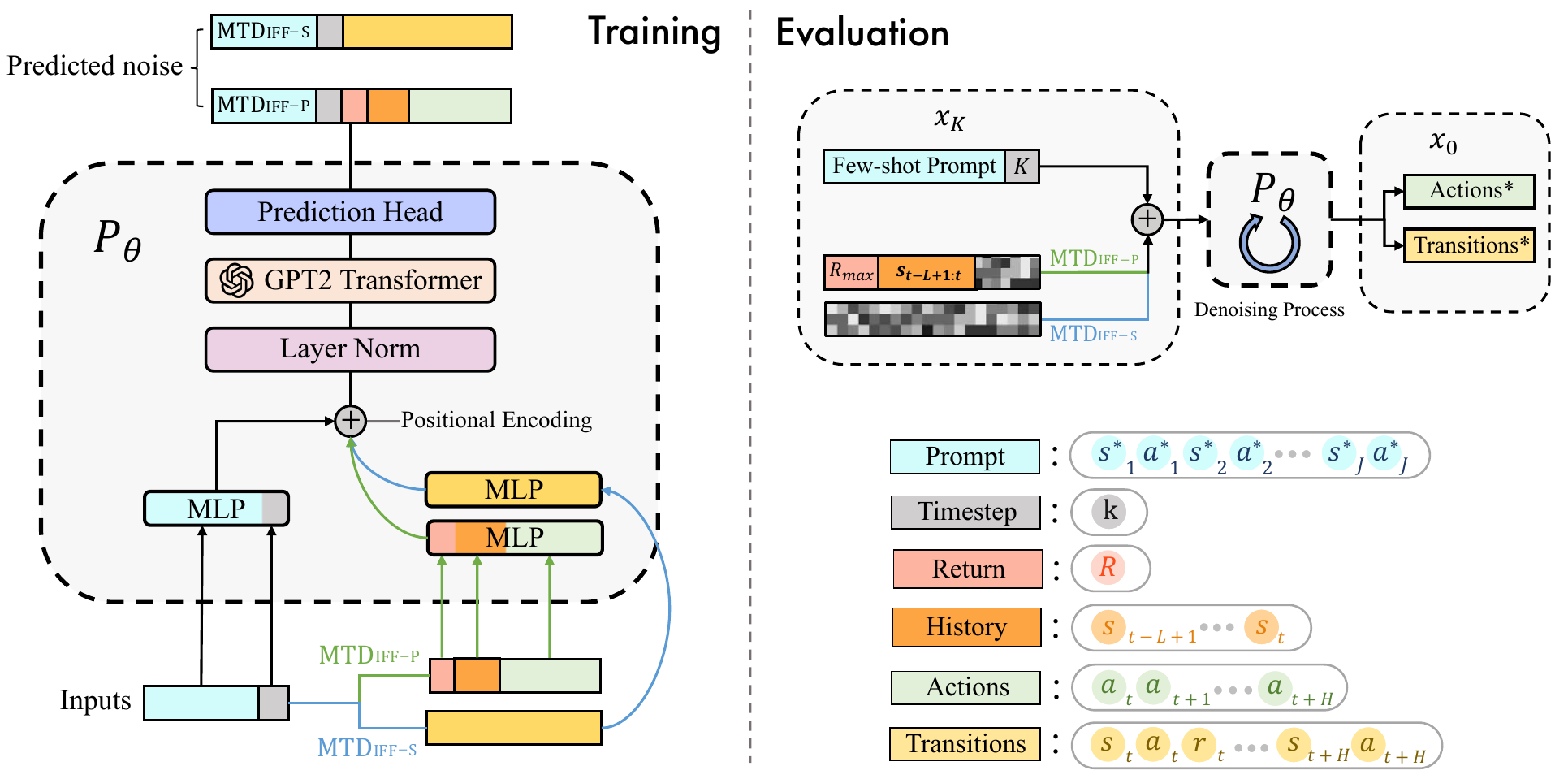}
    \caption{Model architecture of \textsc{MTDiff}, which treats different inputs as tokens in a unified architecture. The two key designs are (\romannumeral 1) the trainable GPT2 Transformer which enhances sequential modeling, and (\romannumeral 2) the MLPs and prediction head which enable efficient training.}
    \label{fig:arc}
\end{figure*}

As shown in Figure~\ref{fig:arc}, first, different raw inputs $x$ are embedded into embeddings $h$ of the same size $\bm{d}$ via separate MLPs $f$ , which can be expressed as:
\begin{equation}
\nonumber
\begin{aligned}
    &h_{P}=f_P(x^{\rm{prompt}}),h_{Ti}=f_{Ti}(x^{\rm{timestep}}),\ \ \ \rhd\:\text{for prompt and diffusion timestep}\\
    &h^{\textcolor{red}{s}}_{Tr}=f_{Tr}(x^{\rm{transitions}}),\:\:\:\:\:\:\rhd\:\text{for\ \textsc{MTDiff-s}}\\
    &h^{\textcolor{red}{p}}_A=f_{A}(x^{\rm{actions}}),h^{\textcolor{red}{p}}_H=f_H(x^{\rm{history}}),
    h^{\textcolor{red}{p}}_R=f_R(x^{\rm{return}}).\:\:\:\:\rhd\:\text{for\ \textsc{MTDiff-p}}
\end{aligned}
\end{equation}
Then, the embeddings $h_P$ and $h_{Ti}$ are prepended as follows to formulate input tokens for \textsc{MTDiff-p} and \textsc{MTDiff-s}, respectively:
\begin{equation}
\nonumber
    h^{\textcolor{red}{p}}_{\rm tokens}={\rm{LN}}(h_{Ti}\times[h_{P},h_{Ti},h_{R}^{\textcolor{red}{p}},h_H^{\textcolor{red}{p}},h_A^{\textcolor{red}{p}}]+ h_R^{\textcolor{red}{p}}+E^{\rm{pos}}),
    h^{\textcolor{red}{s}}_{\rm tokens}={\rm{LN}}(h_{Ti}\times[h_{P},h_{Ti},h_{Tr}^{\textcolor{red}{s}}]+E^{\rm{pos}}),
\end{equation}
where $E^{\rm{pos}}$ is the positional embedding, and $\rm{LN}$ 
 denotes layer normalization \cite{ba2016layer} for stabilizing training. In our implementation, we strengthen the condition of the stacked inputs through multiplication with the diffusion timestep $h_{Ti}$ and addition with the return $h_{R}^{\textcolor{red}{p}}$. 
GPT2 is a decoder-only transformer that incorporates a self-attention mechanism to capture dependencies between different positions in the input sequence. We employ the GPT2 architecture as a trainable backbone in \textsc{MTDiff} to handle sequential inputs. It outputs an updated representation as:
\begin{equation}
\nonumber
    h^{\textcolor{red}{p}}_{\rm out} = {\rm{transformer}}(h^{\textcolor{red}{p}}_{\rm tokens}),\quad h^{\textcolor{red}{s}}_{\rm out} = {\rm{transformer}}(h^{\textcolor{red}{s}}_{\rm tokens}).
\end{equation}
Finally, given the output representation, we use a prediction head consisting of fully connected layers to predict the corresponding noise at diffusion timestep $k$. Notice that the predicted noise shares the same dimensional space as the original inputs, which differs from the representation size $\bm{d}$. This noise is used in the reverse denoising process $p_\theta$ during inference.
We summarize the details of the training process, architecture and hyperparameters used in \textsc{MTDiff} in Appendix~\ref{appendix:mtdiff}. 

\section{Related Work}

\textbf{Diffusion Models in RL.}
Diffusion models have emerged as a powerful family of deep generative models with a record-breaking performance in many applications across vision, language and  combinatorial optimization \cite{imageGen,dalle2,gong2023diffuseq,li2022diffusionlm, li2023t2tco}. Recent works in RL have demonstrated the capability of diffusion models to learn the multimodal distribution of offline policies 
\cite{diffusionql,pearce2023imitating,diffusionpolicy} or human behaviors \cite{behaviormodeling}. Other works formulate the sequential decision-making problem as a conditional generative process \cite{decisiondiffuser} and learn to generate the trajectories satisfying conditioned constraints. However, these works are limited to the single-task settings, while we further study the trajectory modeling and generalization problems of diffusion models in multi-task settings. 

\textbf{Multi-Task RL and Few-Shot RL.}
Multi-task RL aims to learn a shared policy for a diverse set of tasks. The main challenge of multi-task RL is the conflicting gradients among different tasks, and previous online RL works address this problem via gradient surgery \cite{PCG_grad}, conflict-averse learning \cite{liu2021conflict}, and parameter composition \cite{softmodular,sun2022paco}. Instead, \textsc{MTDiff} addresses such a problem in an offline setting through a conditional generative process via a novel transformer architecture. Previous Decision-Transformer (DT)-based methods \cite{gato,gamedt, yu2023using} which consider handling multi-task problems, mainly rely on expert trajectories and entail substantial training expenses. 
Scaled-QL \cite{scaledQL} adopts separate networks for different tasks and is hard to generalize to new tasks. Instead of focusing on the performance of training tasks in multi-task RL, few-shot RL aims to improve the generalizability in novel tasks based on the learned multi-task knowledge. Nevertheless, these methods need additional context encoders \cite{context-1,context-2} or gradient descents in the finetuning stage \cite{sun2023smart,taiga2023investigating,gamedt}. In contrast, we use prompts for few-shot generalization without additional parameter-tuning.

\textbf{Data Augmentation for RL.}
Data augmentation \cite{cobbe2019quantifying,ma2022comprehensive} has been verified to be effective in RL. Previous methods incorporate various data augmentations (e.g. adding noise, random translation) on observations for visual-based RL \cite{drq,laskin2020reinforcement, sinha2022s4rl, laskin2020curl}, which ensure the agents learn on multiple views of the same observation. Differently, we focus on data augmentation via synthesizing new experiences rather than perturbing the origin one. Recent works \cite{yu2023scaling,chen2023genaug} consider augmenting the
observations of robotic control using a text-guided diffusion model whilst maintaining the same action, which differs from our approach that can synthesize novel action and reward labels. The recently proposed SynthER \cite{lu2023synthetic} is closely related to our method by generating transitions of trained tasks via a diffusion model. However, SynthER is studied in single-task settings, while we investigate whether a diffusion model can accommodate all knowledge of multi-task datasets and augment the data for novel tasks.




\section{Experiments}

In this section, we conduct extensive experiments to answer the following questions: (1) How does \textsc{MTDiff-p} compare to other offline and online baselines in the multi-task regime? (2) Does \textsc{MTDiff-s} synthesize high-fidelity data and bring policy improvement? (3) How is \textsc{MTDiff-s} compared with other augmentation methods for single-task RL? (4) Does the synthetic data of \textsc{MTDiff-s} match the original data distribution? (5) Can both \textsc{MTDiff-p} and \textsc{MTDiff-s} generalize to unseen tasks? 

\subsection{Environments and Baselines}

\paragraph{Meta-World Tasks} 
The Meta-World benchmark \cite{metaworld} contains 50 qualitatively-distinct manipulation tasks. The tasks share similar dynamics and require a Sawyer robot to interact with various objects with different shapes, joints, and connectivity. In this setup, the state space and reward functions of different tasks are different since the robot is manipulating different objects with different objectives. At each timestep, the Sawyer robot receives a 4-dimensional fine-grained action, representing the 3D position movements of the end effector and the variation of gripper openness. The original Meta-World environment is configured
with a fixed goal, which is more restrictive and less
realistic in robotic learning. Following recent works \cite{softmodular,sun2022paco}, we extend all the tasks to a random-goal setting and refer to it as MT50-rand. We use the average success rate of all tasks as the evaluation metric. 

By training a SAC \cite{sac} agent for each task in isolation, we utilize the experience collected in the replay buffer as our offline dataset. Similar to \cite{scaledQL}, we consider two different dataset compositions: (\romannumeral 1) \textbf{Near-optimal} dataset consisting of the experience (100M
transitions) from random to expert (convergence) in SAC-Replay, and (\romannumeral 2) \textbf{Sub-optimal} dataset consisting of the initial 50\% of the trajectories (50M transitions) from the replay buffer for each task, where the proportion of expert data decreases a lot. We summarize more details about the dataset in Appendix~\ref{appendix:dataset}.

\paragraph{Maze2D Tasks} Maze2D \cite{d4rl} is a navigation task that requires an agent to traverse from a
randomly designated location to a fixed goal in the 2D map. The reward is 1 if succeed and 0 otherwise. Maze2D can evaluate
the ability of RL algorithms to stitch together previously
collected sub-trajectories, which helps the agent find the shortest path
to evaluation goals. We use the agent's scores as the evaluation metric. The offline dataset is collected by selecting random goal locations and using a planner to generate sequences of waypoints by following a PD controller.


\paragraph{Baselines} We compare our proposed \textsc{MTDiff} (\textsc{MTDiff-p} and \textsc{MTDiff-s}) with the following baselines. Each baseline has the same batch size and training steps as \textsc{MTDiff}. For \textbf{\textsc{MTDiff-p}}, we have following baselines:
(\romannumeral1) \textbf{PromptDT.} PromptDT \cite{promptdt} built on Decision-Transformer (DT) \cite{decisiontransformer} aims to learn from multi-task data and generalize the policy to unseen tasks. PromptDT generates actions based on the trajectory prompts and reward-to-go. We use the same GPT2-network as in \textsc{MTDiff-p}. The main difference between our method and PromptDT is that we employ diffusion models for generative planning.
(\romannumeral2) \textbf{MTDT.} We extend the DT architecture \cite{decisiontransformer} to learn from multi-task data. Specifically, MTDT concatenates an embedding $z$ and a state $s$ as the input tokens, where $z$ is the encoding of task ID. In evaluation, the reward-to-go and task ID are fed into the Transformer to provide task-specific information. MTDT also uses the same GPT2-network as in \textsc{MTDiff-p}. Compared to MTDT, our model incorporates prompt and diffusion framework to learn from the multi-task data.
(\romannumeral3) \textbf{MTCQL.} Following scaled-QL~\cite{scaledQL}, we extend CQL~\cite{cql} with multi-head critic networks and a task-ID conditioned actor for multi-task policy learning. 
(\romannumeral4) \textbf{MTIQL.} We extend IQL \cite{iql} for multi-task learning using a similar revision of MTCQL. The TD-based baselines (i.e., MTCQL and MTIQL) are used to demonstrate the effectiveness of conditional generative modeling for multi-task planning.
(\romannumeral5)~\textbf{MTBC.} We extend Behavior cloning (BC) to multi-task offline policy learning via network scaling and a task-ID conditioned actor that is similar to MTCQL and MTIQL. 


As for \textbf{\textsc{MTDiff-s}}, we compare it with two baselines that perform direct data augmentation in offline RL. 
(\romannumeral1) \textbf{RAD.} We adopt the random amplitude scaling \cite{laskin2020reinforcement} that multiplies a random 
variable to states, i.e., $s' = s \times z$, where $z \sim {\rm Uniform}[\alpha,\beta]$. This augmentation technique has been verified to be effective for state-based RL.
(\romannumeral2) \textbf{S4RL.}  We adopt the adversarial state training \cite{sinha2022s4rl} by taking gradients with respect to the value function to obtain a new state, i.e. $s' \gets s + \epsilon\nabla_s\mathbb{J}_Q(\pi(s))$, where $\mathbb{J}_Q$ is
the policy evaluation update performed via a $Q$ function, and $\epsilon$ is the size of gradient steps. We summarize the details of all the baselines in Appendix~\ref{appendix:algo}.


\subsection{Result Comparison for Planning}

\textbf{How does \textsc{MTDiff-p} compare to baselines in the multi-task regime?} For a fair comparison, 
\begin{wraptable}{r}{.52\textwidth}
\vspace{-1.8em}
\begin{center}
\caption{Average success rate across 3 seeds on Meta-World-V2 MT50 with random goals (MT50-rand). Each task is evaluated for 50 episodes.}
\resizebox{\linewidth}{!}{
\label{Tab:plan-1}
\begin{tabular}{c|cc} 
\toprule[1pt]
\midrule
\textbf{Methods}& \textbf{Near-optimal}&\textbf{Sub-optimal}\\
\midrule
    \textbf{CARE} \cite{care} (Online)&$50.8\pm1.0$&$-$\\
      \textbf{PaCo} \cite{sun2022paco} (Online)&$57.3\pm1.3$&$-$\\
      \midrule
      \textbf{MTDT} & $20.99\pm2.66$&$20.63\pm2.21$\\
      \textbf{PromptDT}&$45.68\pm1.84$&$39.76\pm2.79$\\
      \textbf{MTBC}&$60.39\pm0.86$&$34.53\pm1.25$\\
      \textbf{MTCQL}&$-$&$-$\\
      \textbf{MTIQL}&$56.21\pm1.39$&$43.28\pm0.90$\\
      \midrule
      \textbf{\textsc{MTDiff}-p (ours)}&$59.53\pm1.12$& \cellcolor{red!25}\bm{$48.67\pm1.32$}\\
      \textbf{\textsc{MTDiff-p-onehot} (ours)}&\cellcolor{red!25}\bm{$61.32\pm0.89$}&\cellcolor{red!25}\bm{$48.94\pm0.95$}\\
      \midrule
      \bottomrule[1pt]
    \end{tabular}
    }
  \end{center}
  \end{wraptable}
we add \textsc{MTDiff-p-onehot} as a variant of \textsc{MTDiff-p} by replacing the prompt with a one-hot task-ID, which is used in the baselines except for PromptDT. The first action generated by \textsc{MTDiff-p} is used to interact with the environment. According to Tab.~\ref{Tab:plan-1}, we have the following key observations. (\romannumeral1) Our method achieves better performance than baselines in both near-optimal and sub-optimal settings. For near-optimal datasets, 
\textsc{MTDiff-p} and \textsc{MTDiff-p-onehot} achieve about 60\% success rate, significantly outperforming other methods and performing comparably with MTBC. However, the performance of MTBC decreases a lot in sub-optimal datasets. BC is hard to handle the conflict behaviors in experiences sampled by a mixture of policies with different returns, which has also been verified in previous offline imitation methods \cite{liu2021curriculum,pmlr-v162-xu22l}.
In contrast, both \textsc{MTDiff-p} and \textsc{MTDiff-onehot} perform the best in sub-optimal datasets. (\romannumeral2) We compare \textsc{MTDiff-p} with two SOTA multi-task online RL methods, CARE \cite{care} and PaCo \cite{sun2022paco}, which are trained for 100M steps in MT50-rand. \textsc{MTDiff-p} outperforms both of them given the near-optimal dataset, demonstrating the potential of solving the multi-task RL problems in an offline setting.
(\romannumeral3) MTDT is limited in distinguishing different tasks with task ID while PromptDT performs better, which demonstrates the effect of prompts in multi-task settings. (\romannumeral4) As for TD-based baselines, we find MTCQL almost fails while MTIQL performs well. We hypothesize that since MTCQL penalizes the OOD actions for each task, it will hinder other tasks' learning since different tasks can choose remarkably different actions when facing similar states. In contrast, IQL learns a value function without querying the values of OOD actions.

We remark that \textsc{MTDiff-p} based on GPT network outperforms that with U-Net in similar model size, and detailed results are given in Appendix~\ref{appendix:arch}. Overall, \textsc{MTDiff-p} is an effective planner in multi-task settings
including sub-optimal datasets where we must stitch useful segments of suboptimal trajectories, and near-optimal datasets where we mimic the best behaviors. Meanwhile, we argue that although \textsc{MTDiff-p-onehot} performs well, it cannot generalize to unseen tasks without prompts. 

\begin{figure*}[t]
\centering
\begin{minipage}[h]{0.68\textwidth}
\centering
   \includegraphics[width=1.0\linewidth]{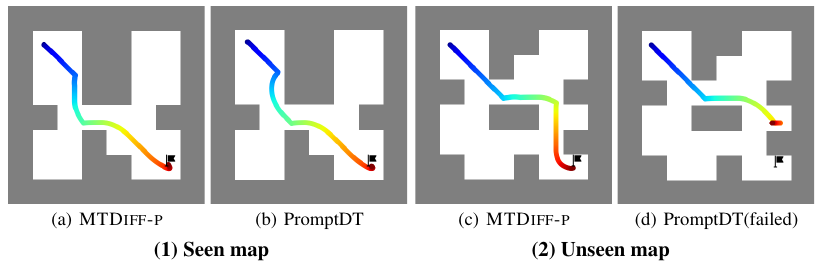}
   \vspace{-2em}
    \caption{Seen and unseen maps of Maze2D with long planning path. Goal position is denoted as {\includegraphics[width=0.015\linewidth]{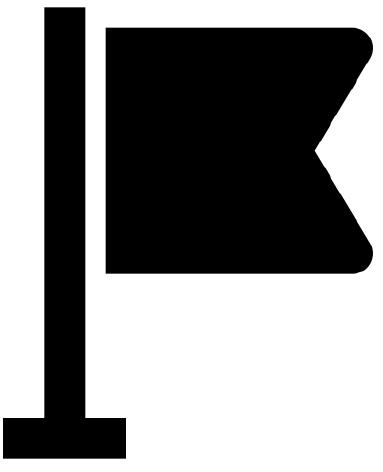}.}
}
\label{fig:maze2d}
\end{minipage}
\hspace{4pt}
\vspace{-1.4em}
\begin{minipage}[h]{0.28\textwidth}
    \centering
    \includegraphics[width=1.0\linewidth]{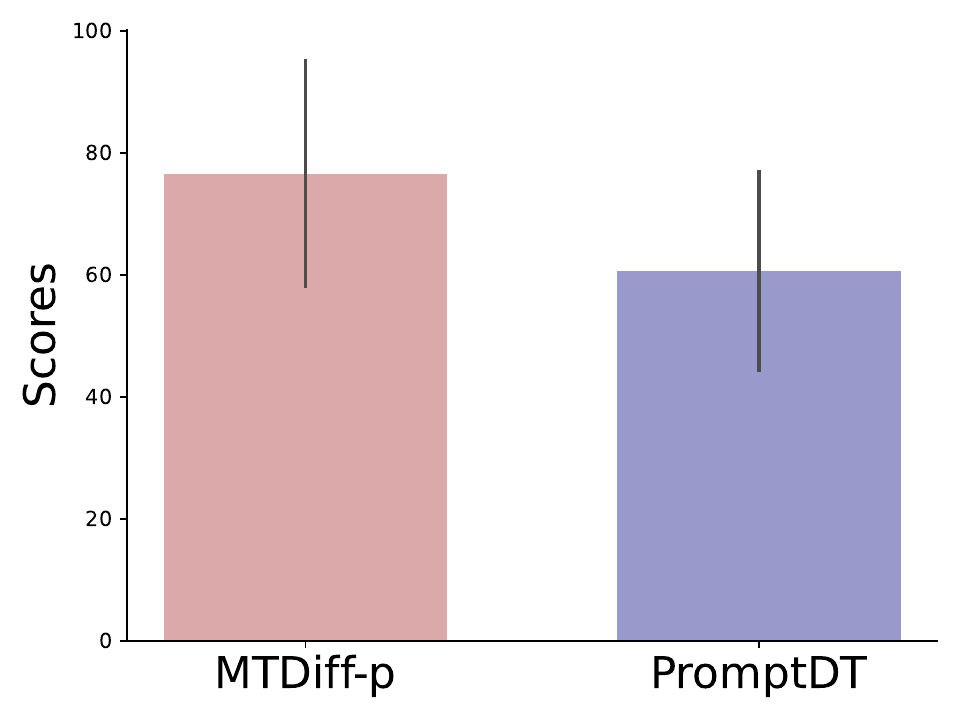}
    \vspace{-2em}
    \caption{Average scores obtained in 8 Maze2D maps.}
    \label{fig:maze_score}
\end{minipage}
\end{figure*}

\paragraph{Does \textsc{MTDiff-p} generalize to unseen tasks?} We further carry out experiments on Maze2D to evaluate the generalizability of \textsc{MTDiff-p}. We select PromptDT as our baseline, as it has demonstrated both competitive performances on training tasks and adaptation ability for unseen tasks \cite{promptdt}. We use 8 different maps for training and one new map for adaptation evaluation. The setup details are given in Appendix~\ref{appendix:maze2d}. We evaluate these two methods on both seen maps and an unseen map. The average scores obtained in 8 training maps are referred to Figure~\ref{fig:maze_score}. To further illustrate the 
advantages of our method compared to PromptDT, we select one difficult training map and the unseen map for visualization, as shown in Figure \ref{fig:maze2d}. 
According to the visualized path, we find that (1) for seen maps in training, \textsc{MTDiff-p} generates a shorter and smoother path, and (2) for unseen maps, PromptDT fails to obtain a reasonable path while \textsc{MTDiff-p} succeed, which verifies that \textsc{MTDiff-p} can perform few-shot adaptation based on trajectory prompts and the designed architecture.

\subsection{Results for Augmentation via Data Synthesis}

\textbf{Does \textsc{MTDiff-s} synthesize high-fidelity data and bring policy improvement?} We train \textsc{MTDiff-s} on near-optimal datasets from 45 tasks to evaluate its generalizability. We select 3 training tasks and 3 unseen tasks, and measure the policy improvement of offline RL training (i.e., TD3-BC \cite{td3bc}) with data augmentation. For each evaluated task, \textsc{MTDiff-s} synthesizes 2M transitions 
to expand the original 1M dataset. From the results summarized in Table \ref{table:augmentation}, \textsc{MTDiff-s} can boost the offline performance for all tasks and significantly increases performance by about 180\%, 131\%, and 161\% for \emph{box-close}, \emph{hand-insert}, and \emph{coffee-push}, respectively.
\begin{figure*}[htp]
\centering
        \vspace{0pt}
    \begin{minipage}[c]{0.52\textwidth}
        \centering
        \includegraphics[width=1\textwidth]{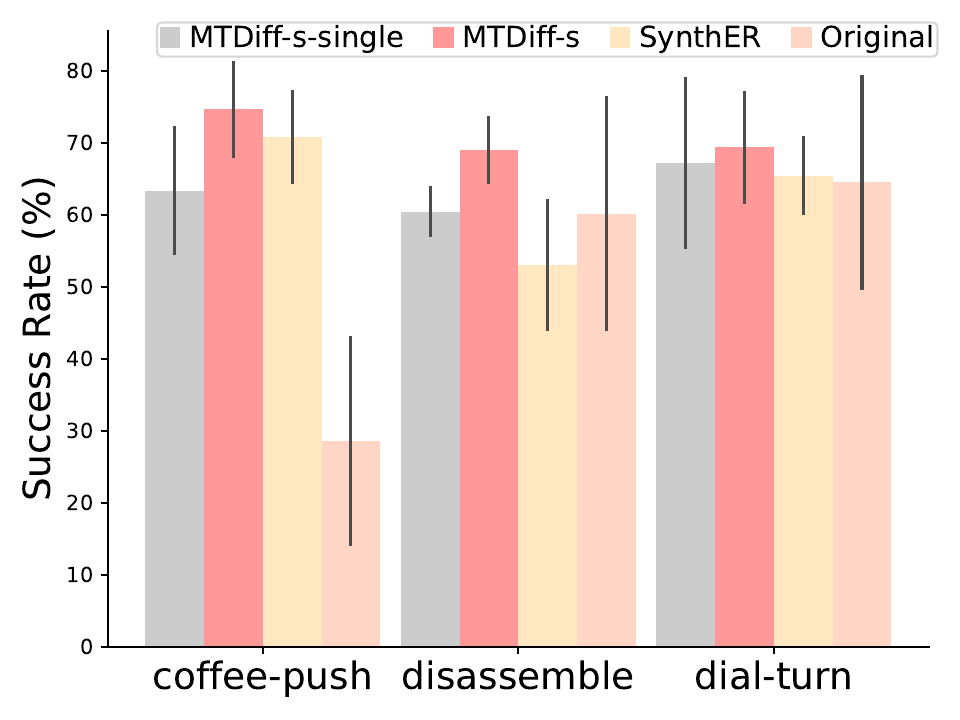}
        \vspace{-15pt}
        \caption{Results of multi-task and single-
task augmentation}
        \label{fig:single_aug}
        \end{minipage}
            \hspace{5pt}        
    \begin{minipage}[c]{0.42\textwidth}
        \includegraphics[width=1\textwidth]{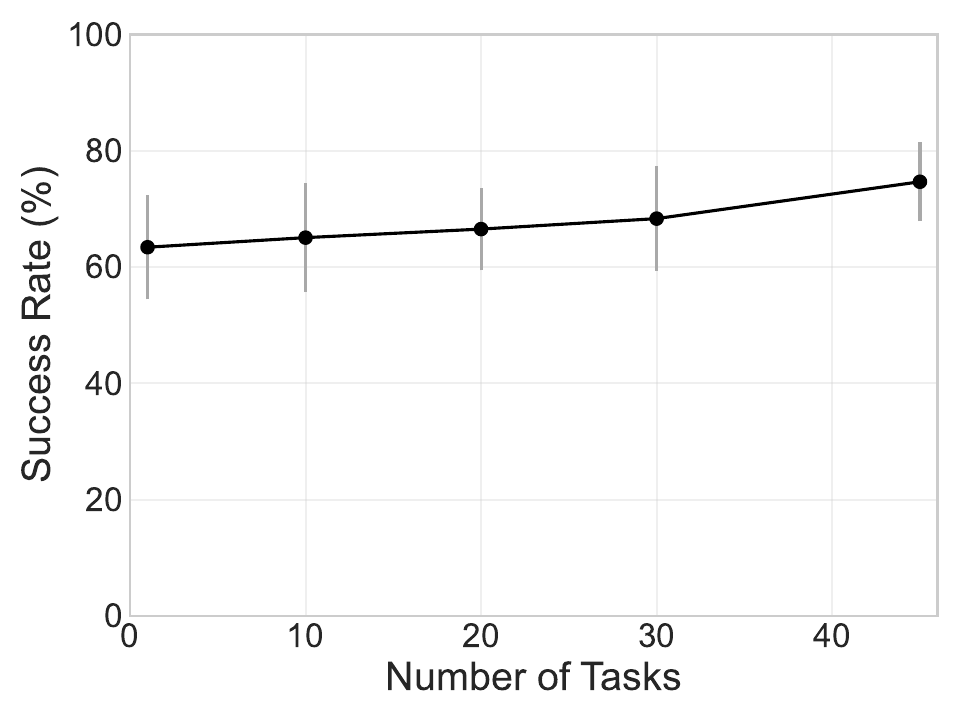}
        \vspace{-15pt}
        \caption{Average success rate across 3 seeds on \emph{coffee-push}. Dataset is augmented by \textsc{MTDiff-s} trained under different number of tasks.}
        \label{fig:num_tasks}
    \end{minipage}
        \vspace{-10pt}
\end{figure*}
\paragraph{How does MTDiff-s perform and benefit from multi-task training?} From Table 
\ref{table:augmentation}, we find \textsc{MTDiff-s} achieves superior policy improvement in seen tasks compared with previous SOTA augmentation methods (i.e., S4RL and RAD) that are developed in single-task RL. We hypothesize that, by absorbing vast knowledge of multi-task data in training, \textsc{MTDiff-s} can perform implicit data sharing \cite{data-share} by integrating other tasks' knowledge into data synthesis of the current task. To verify this hypothesis, we select three tasks (i.e., \emph{coffee-push}, \emph{disassemble} and \emph{dial-turn}) to re-train \textsc{MTDiff-s} on the corresponding single-task dataset. 
We denote this variant as \textsc{MTDiff-s-single}. We implement SynthER \cite{lu2023synthetic} to further confirm the benefits of multi-task training. SynthER also utilizes diffusion models to enhance offline datasets for policy improvement, but it doesn't take into account the aspect of learning from multi-task data. We observe that \textsc{MTDiff-s} outperforms both two methods, as shown in Figure~\ref{fig:single_aug}. What's more, we re-train \textsc{MTDiff-s} on 10, 20, and 30 tasks respectively, in order to obtain the relationship between the performance gains and the number of training tasks. Our findings, as outlined in Fig.~\ref{fig:num_tasks}, provide compelling evidence in support of our hypothesis: \textsc{MTDiff-s} exhibits progressively superior data synthesis performance with increasing task diversity.
\paragraph{Does \textsc{MTDiff-s} generalize to unseen tasks?}
We answer this question by conducting offline RL training on the augmented datasets of 3 unseen tasks. According to Table \ref{table:augmentation}, \textsc{MTDiff-s} is well-generalized and obtains significant improvement compared to the success rate of original datasets. \textsc{MTDiff-s} boost the policy performance by 131\%, 180\% and 32\% for \emph{hand-insert}, \emph{box-close} and \emph{bin-picking}, respectively. We remark that S4RL performs the best on the two unseen tasks, i.e., \emph{box-close} and \emph{bin-picking}, since it utilizes the entire datasets to train $Q$-functions and obtains the augmented states. Nevertheless, we use much less information (i.e., a single trajectory as prompts) for augmentation.

\begin{table}[htbp]
\begin{center}
\caption{Average success rate across 3 seeds on Meta-World-V2 single task with random goals. Each selected task is evaluated for 500 episodes.}
\vspace{-1.0em}
\label{table:augmentation}
\resizebox{\linewidth}{!}{
\begin{tabular}{cccccc} 
    \hline
    \multicolumn{2}{c}{\textbf{Tasks}}&\textbf{Original}&\textbf{S4RL}&\textbf{RAD}&\textbf{\textsc{MTDiff-s(ours)}}\\
    \hline
    \multirow{3}{*}{\textbf{Unseen Tasks}}& \textbf{box-close} & $23.46\pm7.11$&$\cellcolor{red!25}\bm{73.13\pm3.51}$\ \color{blue}($\uparrow 211\%)$&$71.20\pm4.84$\ \color{blue}$(\uparrow 203\%)$& $65.73\pm8.36$\ \color{blue}$(\uparrow 180\%)$\\
    &\textbf{hand-insert}&$30.60\pm9.77$&$60.20\pm1.57$\ \color{blue}$(\uparrow 96\%)$&$43.79\pm3.44$\ \color{blue}$(\uparrow 43\%)$&$\cellcolor{red!25}\bm{70.87\pm3.59}$\ \color{blue}$(\uparrow 131\%)$\\
    &\textbf{bin-picking}&$42.13\pm14.33$&$\cellcolor{red!25}\bm{72.20\pm4.17}$\color{blue}$(\uparrow 71\%)$&$43.27\pm4.38$\color{blue}$(\uparrow 2\%)$&$55.73\pm7.63$\color{blue}$(\uparrow 32\%)$\\
    \hline
    \multirow{3}{*}{\textbf{Seen Tasks}}&\textbf{sweep-into}&$91.8\pm1.14$&$90.53\pm3.52$\color{blue}$(\downarrow 1\%)$&$88.06\pm9.86$\color{blue}$(\downarrow 4\%)$&$\cellcolor{red!25}\bm{92.87\pm1.11}$\color{blue}$(\uparrow 1\%)$\\
    &\textbf{coffee-push}&$28.60\pm14.55$&$28.73\pm8.44$\color{blue}$(\uparrow 0.4\%)$&$33.19\pm2.86$\color{blue}$(\uparrow 16\%)$&$\cellcolor{red!25}\bm{74.67\pm6.79}$\color{blue}$(\uparrow 161\%)$\\
    &\textbf{disassemble}&$60.20\pm16.29$&$52.20\pm5.68$\color{blue}$(\downarrow 12\%)$&$60.93\pm20.80$\color{blue}$(\uparrow 1\%)$&$\cellcolor{red!25}\bm{69.00\pm4.72}$\color{blue}$(\uparrow 14.6\%)$\\
      \hline
    \end{tabular}
    }
\vspace{-1em}
\end{center}
\end{table}

\textbf{Does the synthetic data of \textsc{MTDiff-s} match the original data distribution?}
We select 4 tasks and use T-SNE~\cite{tsne} to visualize the distribution of original data and synthetic data. We find the synthetic data overlap and expand the original data distribution while also keeping consistency with the underlying MDP. 
The visualization results and further analyses are given in Appendix~\ref{appendix:vis}.

\vspace{-0.5em}
\section{Conclusion}
\vspace{-0.5em}

We propose \textsc{MTDiff}, a diffusion-based effective planner and data synthesizer for multi-task RL. 
With the trajectory prompt and unified GPT-based architecture, \textsc{MTDiff} can model multi-task data and generalize to unseen tasks. We show that in the MT50-rand benchmark containing fine-grained manipulation tasks, \textsc{MTDiff-p} generates desirable behavior for each task via few-shot prompts. By compressing multi-task knowledge in a single model, we demonstrate that \textsc{MTDiff-s} greatly boosts policy performance by augmenting original offline datasets. In future research, 
we aim to develop a practical multi-task algorithm for real robots to trade off the sample speed and generative quality. We further discuss the limitations and broader impacts of \textsc{MTDiff} in Appendix~\ref{appendix:limi}. 
\section*{Acknowledgments}
This work is supported by the National Natural Science Foundation of China (Grant No.62306242\&62076161), the National Key R\&D Program of China (Grant No.2022ZD0160100), Shanghai Municipal Science and Technology Major Project (2021SHZDZX0102) and Shanghai Artificial Intelligence Laboratory.
\bibliographystyle{plainnat}
\bibliography{nips}

\begin{thebibliography}{79}
\providecommand{\natexlab}[1]{#1}
\providecommand{\url}[1]{\texttt{#1}}
\expandafter\ifx\csname urlstyle\endcsname\relax
  \providecommand{\doi}[1]{doi: #1}\else
  \providecommand{\doi}{doi: \begingroup \urlstyle{rm}\Url}\fi

\bibitem[Ahn et~al.(2022)Ahn, Brohan, Brown, Chebotar, Cortes, David, Finn,
  Gopalakrishnan, Hausman, Herzog, et~al.]{saycan}
Michael Ahn, Anthony Brohan, Noah Brown, Yevgen Chebotar, Omar Cortes, Byron
  David, Chelsea Finn, Keerthana Gopalakrishnan, Karol Hausman, Alex Herzog,
  et~al.
\newblock Do as i can, not as i say: Grounding language in robotic affordances.
\newblock \emph{arXiv preprint arXiv:2204.01691}, 2022.

\bibitem[Ajay et~al.(2023)Ajay, Du, Gupta, Tenenbaum, Jaakkola, and
  Agrawal]{decisiondiffuser}
Anurag Ajay, Yilun Du, Abhi Gupta, Joshua~B. Tenenbaum, Tommi~S. Jaakkola, and
  Pulkit Agrawal.
\newblock Is conditional generative modeling all you need for decision making?
\newblock In \emph{The Eleventh International Conference on Learning
  Representations}, 2023.

\bibitem[Ba et~al.(2016)Ba, Kiros, and Hinton]{ba2016layer}
Lei~Jimmy Ba, Jamie~Ryan Kiros, and Geoffrey~E. Hinton.
\newblock Layer normalization.
\newblock \emph{CoRR}, abs/1607.06450, 2016.
\newblock URL \url{http://arxiv.org/abs/1607.06450}.

\bibitem[Bai et~al.(2018)Bai, Kolter, and Koltun]{tcn}
Shaojie Bai, J~Zico Kolter, and Vladlen Koltun.
\newblock An empirical evaluation of generic convolutional and recurrent
  networks for sequence modeling.
\newblock \emph{arXiv preprint arXiv:1803.01271}, 2018.

\bibitem[Bao et~al.(2023)Bao, Nie, Xue, Li, Pu, Wang, Yue, Cao, Su, and
  Zhu]{bao2023transformer}
Fan Bao, Shen Nie, Kaiwen Xue, Chongxuan Li, Shi Pu, Yaole Wang, Gang Yue, Yue
  Cao, Hang Su, and Jun Zhu.
\newblock One transformer fits all distributions in multi-modal diffusion at
  scale.
\newblock \emph{arXiv preprint arXiv:2303.06555}, 2023.

\bibitem[Brohan et~al.(2022)Brohan, Brown, Carbajal, Chebotar, Dabis, Finn,
  Gopalakrishnan, Hausman, Herzog, Hsu, et~al.]{brohan2022rt1}
Anthony Brohan, Noah Brown, Justice Carbajal, Yevgen Chebotar, Joseph Dabis,
  Chelsea Finn, Keerthana Gopalakrishnan, Karol Hausman, Alex Herzog, Jasmine
  Hsu, et~al.
\newblock Rt-1: Robotics transformer for real-world control at scale.
\newblock \emph{arXiv preprint arXiv:2212.06817}, 2022.

\bibitem[Brown et~al.(2020)Brown, Mann, Ryder, Subbiah, Kaplan, Dhariwal,
  Neelakantan, Shyam, Sastry, Askell, et~al.]{gpt3}
Tom Brown, Benjamin Mann, Nick Ryder, Melanie Subbiah, Jared~D Kaplan, Prafulla
  Dhariwal, Arvind Neelakantan, Pranav Shyam, Girish Sastry, Amanda Askell,
  et~al.
\newblock Language models are few-shot learners.
\newblock \emph{Advances in neural information processing systems},
  33:\penalty0 1877--1901, 2020.

\bibitem[Chen et~al.(2023{\natexlab{a}})Chen, Lu, Ying, Su, and
  Zhu]{behaviormodeling}
Huayu Chen, Cheng Lu, Chengyang Ying, Hang Su, and Jun Zhu.
\newblock Offline reinforcement learning via high-fidelity generative behavior
  modeling.
\newblock In \emph{The Eleventh International Conference on Learning
  Representations}, 2023{\natexlab{a}}.

\bibitem[Chen et~al.(2021)Chen, Lu, Rajeswaran, Lee, Grover, Laskin, Abbeel,
  Srinivas, and Mordatch]{decisiontransformer}
Lili Chen, Kevin Lu, Aravind Rajeswaran, Kimin Lee, Aditya Grover, Michael
  Laskin, Pieter Abbeel, Aravind Srinivas, and Igor Mordatch.
\newblock Decision transformer: Reinforcement learning via sequence modeling.
\newblock In A.~Beygelzimer, Y.~Dauphin, P.~Liang, and J.~Wortman Vaughan,
  editors, \emph{Advances in Neural Information Processing Systems}, 2021.
\newblock URL \url{https://openreview.net/forum?id=a7APmM4B9d}.

\bibitem[Chen et~al.(2023{\natexlab{b}})Chen, Kiami, Gupta, and
  Kumar]{chen2023genaug}
Zoey Chen, Sho Kiami, Abhishek Gupta, and Vikash Kumar.
\newblock Genaug: Retargeting behaviors to unseen situations via generative
  augmentation.
\newblock \emph{arXiv preprint arXiv:2302.06671}, 2023{\natexlab{b}}.

\bibitem[Chi et~al.(2023)Chi, Feng, Du, Xu, Cousineau, Burchfiel, and
  Song]{diffusionpolicy}
Cheng Chi, Siyuan Feng, Yilun Du, Zhenjia Xu, Eric Cousineau, Benjamin
  Burchfiel, and Shuran Song.
\newblock Diffusion policy: Visuomotor policy learning via action diffusion.
\newblock \emph{arXiv preprint arXiv:2303.04137}, 2023.

\bibitem[Chowdhery et~al.(2022)Chowdhery, Narang, Devlin, Bosma, Mishra,
  Roberts, Barham, Chung, Sutton, Gehrmann, et~al.]{chowdhery2022palm}
Aakanksha Chowdhery, Sharan Narang, Jacob Devlin, Maarten Bosma, Gaurav Mishra,
  Adam Roberts, Paul Barham, Hyung~Won Chung, Charles Sutton, Sebastian
  Gehrmann, et~al.
\newblock Palm: Scaling language modeling with pathways.
\newblock \emph{arXiv preprint arXiv:2204.02311}, 2022.

\bibitem[Cobbe et~al.(2019)Cobbe, Klimov, Hesse, Kim, and
  Schulman]{cobbe2019quantifying}
Karl Cobbe, Oleg Klimov, Chris Hesse, Taehoon Kim, and John Schulman.
\newblock Quantifying generalization in reinforcement learning.
\newblock In \emph{International Conference on Machine Learning}, pages
  1282--1289. PMLR, 2019.

\bibitem[Ebert et~al.(2021)Ebert, Yang, Schmeckpeper, Bucher, Georgakis,
  Daniilidis, Finn, and Levine]{bridge_data}
Frederik Ebert, Yanlai Yang, Karl Schmeckpeper, Bernadette Bucher, Georgios
  Georgakis, Kostas Daniilidis, Chelsea Finn, and Sergey Levine.
\newblock Bridge data: Boosting generalization of robotic skills with
  cross-domain datasets.
\newblock \emph{CoRR}, abs/2109.13396, 2021.
\newblock URL \url{https://arxiv.org/abs/2109.13396}.

\bibitem[Fu et~al.(2020)Fu, Kumar, Nachum, Tucker, and Levine]{d4rl}
Justin Fu, Aviral Kumar, Ofir Nachum, George Tucker, and Sergey Levine.
\newblock {D4RL:} datasets for deep data-driven reinforcement learning.
\newblock \emph{CoRR}, abs/2004.07219, 2020.
\newblock URL \url{https://arxiv.org/abs/2004.07219}.

\bibitem[Fujimoto and Gu(2021)]{td3bc}
Scott Fujimoto and Shixiang~Shane Gu.
\newblock A minimalist approach to offline reinforcement learning.
\newblock In \emph{Thirty-Fifth Conference on Neural Information Processing
  Systems}, 2021.

\bibitem[Gong et~al.(2023)Gong, Li, Feng, Wu, and Kong]{gong2023diffuseq}
Shansan Gong, Mukai Li, Jiangtao Feng, Zhiyong Wu, and Lingpeng Kong.
\newblock Diffuseq: Sequence to sequence text generation with diffusion models.
\newblock In \emph{The Eleventh International Conference on Learning
  Representations}, 2023.
\newblock URL \url{https://openreview.net/forum?id=jQj-_rLVXsj}.

\bibitem[Haarnoja et~al.(2018)Haarnoja, Zhou, Abbeel, and Levine]{sac}
Tuomas Haarnoja, Aurick Zhou, Pieter Abbeel, and Sergey Levine.
\newblock Soft actor-critic: Off-policy maximum entropy deep reinforcement
  learning with a stochastic actor.
\newblock In \emph{International conference on machine learning}, pages
  1861--1870. PMLR, 2018.

\bibitem[Ho and Salimans(2021)]{ho2021classifierfree}
Jonathan Ho and Tim Salimans.
\newblock Classifier-free diffusion guidance.
\newblock In \emph{NeurIPS 2021 Workshop on Deep Generative Models and
  Downstream Applications}, 2021.
\newblock URL \url{https://openreview.net/forum?id=qw8AKxfYbI}.

\bibitem[Ho et~al.(2020)Ho, Jain, and Abbeel]{ddpm}
Jonathan Ho, Ajay Jain, and Pieter Abbeel.
\newblock Denoising diffusion probabilistic models.
\newblock \emph{Advances in Neural Information Processing Systems},
  33:\penalty0 6840--6851, 2020.

\bibitem[Janner et~al.(2022)Janner, Du, Tenenbaum, and Levine]{diffuser}
Michael Janner, Yilun Du, Joshua~B. Tenenbaum, and Sergey Levine.
\newblock Planning with diffusion for flexible behavior synthesis.
\newblock In \emph{International Conference on Machine Learning}, 2022.

\bibitem[Kalashnikov et~al.(2021)Kalashnikov, Varley, Chebotar, Swanson,
  Jonschkowski, Finn, Levine, and Hausman]{mt-opt}
Dmitry Kalashnikov, Jacob Varley, Yevgen Chebotar, Benjamin Swanson, Rico
  Jonschkowski, Chelsea Finn, Sergey Levine, and Karol Hausman.
\newblock Mt-opt: Continuous multi-task robotic reinforcement learning at
  scale.
\newblock \emph{CoRR}, abs/2104.08212, 2021.
\newblock URL \url{https://arxiv.org/abs/2104.08212}.

\bibitem[Kingma and Ba(2014)]{kingma2014adam}
Diederik~P Kingma and Jimmy Ba.
\newblock Adam: A method for stochastic optimization.
\newblock \emph{arXiv preprint arXiv:1412.6980}, 2014.

\bibitem[Kostrikov et~al.(2022)Kostrikov, Nair, and Levine]{iql}
Ilya Kostrikov, Ashvin Nair, and Sergey Levine.
\newblock Offline reinforcement learning with implicit q-learning.
\newblock In \emph{International Conference on Learning Representations}, 2022.
\newblock URL \url{https://openreview.net/forum?id=68n2s9ZJWF8}.

\bibitem[Kumar et~al.(2020)Kumar, Zhou, Tucker, and Levine]{cql}
Aviral Kumar, Aurick Zhou, George Tucker, and Sergey Levine.
\newblock Conservative q-learning for offline reinforcement learning.
\newblock \emph{CoRR}, abs/2006.04779, 2020.
\newblock URL \url{https://arxiv.org/abs/2006.04779}.

\bibitem[Kumar et~al.(2023)Kumar, Agarwal, Geng, Tucker, and Levine]{scaledQL}
Aviral Kumar, Rishabh Agarwal, Xinyang Geng, George Tucker, and Sergey Levine.
\newblock Offline q-learning on diverse multi-task data both scales and
  generalizes.
\newblock In \emph{The Eleventh International Conference on Learning
  Representations}, 2023.

\bibitem[Laskin et~al.(2020{\natexlab{a}})Laskin, Srinivas, and
  Abbeel]{laskin2020curl}
Michael Laskin, Aravind Srinivas, and Pieter Abbeel.
\newblock Curl: Contrastive unsupervised representations for reinforcement
  learning.
\newblock In \emph{International Conference on Machine Learning}, pages
  5639--5650. PMLR, 2020{\natexlab{a}}.

\bibitem[Laskin et~al.(2020{\natexlab{b}})Laskin, Lee, Stooke, Pinto, Abbeel,
  and Srinivas]{laskin2020reinforcement}
Misha Laskin, Kimin Lee, Adam Stooke, Lerrel Pinto, Pieter Abbeel, and Aravind
  Srinivas.
\newblock Reinforcement learning with augmented data.
\newblock \emph{Advances in neural information processing systems},
  33:\penalty0 19884--19895, 2020{\natexlab{b}}.

\bibitem[Lee et~al.(2022)Lee, Nachum, Yang, Lee, Freeman, Guadarrama, Fischer,
  Xu, Jang, Michalewski, et~al.]{gamedt}
Kuang-Huei Lee, Ofir Nachum, Mengjiao~Sherry Yang, Lisa Lee, Daniel Freeman,
  Sergio Guadarrama, Ian Fischer, Winnie Xu, Eric Jang, Henryk Michalewski,
  et~al.
\newblock Multi-game decision transformers.
\newblock \emph{Advances in Neural Information Processing Systems},
  35:\penalty0 27921--27936, 2022.

\bibitem[Levine et~al.(2020)Levine, Kumar, Tucker, and Fu]{offlinerl}
Sergey Levine, Aviral Kumar, George Tucker, and Justin Fu.
\newblock Offline reinforcement learning: Tutorial, review, and perspectives on
  open problems.
\newblock \emph{CoRR}, abs/2005.01643, 2020.
\newblock URL \url{https://arxiv.org/abs/2005.01643}.

\bibitem[Li et~al.(2022)Li, Thickstun, Gulrajani, Liang, and
  Hashimoto]{li2022diffusionlm}
Xiang~Lisa Li, John Thickstun, Ishaan Gulrajani, Percy Liang, and Tatsunori
  Hashimoto.
\newblock Diffusion-{LM} improves controllable text generation.
\newblock In Alice~H. Oh, Alekh Agarwal, Danielle Belgrave, and Kyunghyun Cho,
  editors, \emph{Advances in Neural Information Processing Systems}, 2022.
\newblock URL \url{https://openreview.net/forum?id=3s9IrEsjLyk}.

\bibitem[Li et~al.(2023)Li, Guo, Wang, and Yan]{li2023t2tco}
Yang Li, Jinpei Guo, Runzhong Wang, and Junchi Yan.
\newblock From distribution learning in training to gradient search in testing
  for combinatorial optimization.
\newblock In \emph{Advances in Neural Information Processing Systems}, 2023.

\bibitem[Liang et~al.(2023)Liang, Mu, Ding, Ni, Tomizuka, and
  Luo]{AdaptDiffuser}
Zhixuan Liang, Yao Mu, Mingyu Ding, Fei Ni, Masayoshi Tomizuka, and Ping Luo.
\newblock Adaptdiffuser: Diffusion models as adaptive self-evolving planners.
\newblock \emph{ArXiv}, abs/2302.01877, 2023.

\bibitem[Liu et~al.(2021{\natexlab{a}})Liu, Liu, Jin, Stone, and
  Liu]{liu2021conflict}
Bo~Liu, Xingchao Liu, Xiaojie Jin, Peter Stone, and Qiang Liu.
\newblock Conflict-averse gradient descent for multi-task learning.
\newblock \emph{Advances in Neural Information Processing Systems},
  34:\penalty0 18878--18890, 2021{\natexlab{a}}.

\bibitem[Liu et~al.(2021{\natexlab{b}})Liu, Zhao, Yang, Shen, Zhang, Zhao, and
  Liu]{liu2021curriculum}
Minghuan Liu, Hanye Zhao, Zhengyu Yang, Jian Shen, Weinan Zhang, Li~Zhao, and
  Tie-Yan Liu.
\newblock Curriculum offline imitating learning.
\newblock In A.~Beygelzimer, Y.~Dauphin, P.~Liang, and J.~Wortman Vaughan,
  editors, \emph{Advances in Neural Information Processing Systems},
  2021{\natexlab{b}}.
\newblock URL \url{https://openreview.net/forum?id=q6Kknb68dQf}.

\bibitem[Lu et~al.(2022{\natexlab{a}})Lu, Zhou, Bao, Chen, Li, and
  Zhu]{lu2022dpm++}
Cheng Lu, Yuhao Zhou, Fan Bao, Jianfei Chen, Chongxuan Li, and Jun Zhu.
\newblock Dpm-solver++: Fast solver for guided sampling of diffusion
  probabilistic models.
\newblock \emph{arXiv preprint arXiv:2211.01095}, 2022{\natexlab{a}}.

\bibitem[Lu et~al.(2022{\natexlab{b}})Lu, Zhou, Bao, Chen, Li, and
  Zhu]{lu2022dpmsolver}
Cheng Lu, Yuhao Zhou, Fan Bao, Jianfei Chen, Chongxuan Li, and Jun Zhu.
\newblock {DPM}-solver: A fast {ODE} solver for diffusion probabilistic model
  sampling in around 10 steps.
\newblock In Alice~H. Oh, Alekh Agarwal, Danielle Belgrave, and Kyunghyun Cho,
  editors, \emph{Advances in Neural Information Processing Systems},
  2022{\natexlab{b}}.

\bibitem[Lu et~al.(2023)Lu, Ball, and Parker-Holder]{lu2023synthetic}
Cong Lu, Philip~J. Ball, and Jack Parker-Holder.
\newblock Synthetic experience replay.
\newblock In \emph{Workshop on Reincarnating Reinforcement Learning at ICLR
  2023}, 2023.
\newblock URL \url{https://openreview.net/forum?id=0a9p3Ty2k_}.

\bibitem[Ma et~al.(2022)Ma, Wang, Yuan, Wang, Yuan, and
  Tao]{ma2022comprehensive}
Guozheng Ma, Zhen Wang, Zhecheng Yuan, Xueqian Wang, Bo~Yuan, and Dacheng Tao.
\newblock A comprehensive survey of data augmentation in visual reinforcement
  learning.
\newblock \emph{arXiv preprint arXiv:2210.04561}, 2022.

\bibitem[Mees et~al.(2022{\natexlab{a}})Mees, Hermann, and Burgard]{LCR_IL}
Oier Mees, Lukas Hermann, and Wolfram Burgard.
\newblock What matters in language conditioned robotic imitation learning over
  unstructured data.
\newblock \emph{IEEE Robotics and Automation Letters}, 7\penalty0 (4):\penalty0
  11205--11212, 2022{\natexlab{a}}.
\newblock \doi{10.1109/LRA.2022.3196123}.

\bibitem[Mees et~al.(2022{\natexlab{b}})Mees, Hermann, Rosete-Beas, and
  Burgard]{calvin}
Oier Mees, Lukas Hermann, Erick Rosete-Beas, and Wolfram Burgard.
\newblock Calvin: A benchmark for language-conditioned policy learning for
  long-horizon robot manipulation tasks.
\newblock \emph{IEEE Robotics and Automation Letters}, 7\penalty0 (3):\penalty0
  7327--7334, 2022{\natexlab{b}}.
\newblock \doi{10.1109/LRA.2022.3180108}.

\bibitem[OpenAI(2023)]{openai2023gpt4}
OpenAI.
\newblock Gpt-4 technical report, 2023.

\bibitem[Pearce et~al.(2023)Pearce, Rashid, Kanervisto, Bignell, Sun,
  Georgescu, Macua, Tan, Momennejad, Hofmann, and Devlin]{pearce2023imitating}
Tim Pearce, Tabish Rashid, Anssi Kanervisto, Dave Bignell, Mingfei Sun, Raluca
  Georgescu, Sergio~Valcarcel Macua, Shan~Zheng Tan, Ida Momennejad, Katja
  Hofmann, and Sam Devlin.
\newblock Imitating human behaviour with diffusion models.
\newblock In \emph{The Eleventh International Conference on Learning
  Representations}, 2023.

\bibitem[Peebles and Xie(2022)]{Peebles2022DiT}
William Peebles and Saining Xie.
\newblock Scalable diffusion models with transformers.
\newblock \emph{arXiv preprint arXiv:2212.09748}, 2022.

\bibitem[Radford et~al.(2019)Radford, Wu, Child, Luan, Amodei, Sutskever,
  et~al.]{gpt2}
Alec Radford, Jeffrey Wu, Rewon Child, David Luan, Dario Amodei, Ilya
  Sutskever, et~al.
\newblock Language models are unsupervised multitask learners.
\newblock \emph{OpenAI blog}, 1\penalty0 (8):\penalty0 9, 2019.

\bibitem[Ramesh et~al.(2022)Ramesh, Dhariwal, Nichol, Chu, and Chen]{dalle2}
Aditya Ramesh, Prafulla Dhariwal, Alex Nichol, Casey Chu, and Mark Chen.
\newblock Hierarchical text-conditional image generation with clip latents,
  2022.

\bibitem[Reed et~al.(2022)Reed, Zolna, Parisotto, Colmenarejo, Novikov,
  Barth-Maron, Gimenez, Sulsky, Kay, Springenberg, et~al.]{gato}
Scott Reed, Konrad Zolna, Emilio Parisotto, Sergio~Gomez Colmenarejo, Alexander
  Novikov, Gabriel Barth-Maron, Mai Gimenez, Yury Sulsky, Jackie Kay,
  Jost~Tobias Springenberg, et~al.
\newblock A generalist agent.
\newblock \emph{arXiv preprint arXiv:2205.06175}, 2022.

\bibitem[Ronneberger et~al.(2015)Ronneberger, Fischer, and Brox]{unet}
Olaf Ronneberger, Philipp Fischer, and Thomas Brox.
\newblock U-net: Convolutional networks for biomedical image segmentation.
\newblock In \emph{Medical Image Computing and Computer-Assisted
  Intervention--MICCAI 2015: 18th International Conference, Munich, Germany,
  October 5-9, 2015, Proceedings, Part III 18}, pages 234--241. Springer, 2015.

\bibitem[Saharia et~al.(2022)Saharia, Chan, Saxena, Li, Whang, Denton,
  Ghasemipour, Ayan, Mahdavi, Lopes, Salimans, Ho, Fleet, and
  Norouzi]{imageGen}
Chitwan Saharia, William Chan, Saurabh Saxena, Lala Li, Jay Whang, Emily
  Denton, Seyed Kamyar~Seyed Ghasemipour, Burcu~Karagol Ayan, S.~Sara Mahdavi,
  Rapha~Gontijo Lopes, Tim Salimans, Jonathan Ho, David~J Fleet, and Mohammad
  Norouzi.
\newblock Photorealistic text-to-image diffusion models with deep language
  understanding, 2022.

\bibitem[Sanh et~al.(2022)Sanh, Webson, Raffel, Bach, Sutawika, Alyafeai,
  Chaffin, Stiegler, Raja, Dey, Bari, Xu, Thakker, Sharma, Szczechla, Kim,
  Chhablani, Nayak, Datta, Chang, Jiang, Wang, Manica, Shen, Yong, Pandey,
  Bawden, Wang, Neeraj, Rozen, Sharma, Santilli, Fevry, Fries, Teehan, Scao,
  Biderman, Gao, Wolf, and Rush]{prompt4zeroshot}
Victor Sanh, Albert Webson, Colin Raffel, Stephen Bach, Lintang Sutawika, Zaid
  Alyafeai, Antoine Chaffin, Arnaud Stiegler, Arun Raja, Manan Dey, M~Saiful
  Bari, Canwen Xu, Urmish Thakker, Shanya~Sharma Sharma, Eliza Szczechla,
  Taewoon Kim, Gunjan Chhablani, Nihal Nayak, Debajyoti Datta, Jonathan Chang,
  Mike Tian-Jian Jiang, Han Wang, Matteo Manica, Sheng Shen, Zheng~Xin Yong,
  Harshit Pandey, Rachel Bawden, Thomas Wang, Trishala Neeraj, Jos Rozen,
  Abheesht Sharma, Andrea Santilli, Thibault Fevry, Jason~Alan Fries, Ryan
  Teehan, Teven~Le Scao, Stella Biderman, Leo Gao, Thomas Wolf, and Alexander~M
  Rush.
\newblock Multitask prompted training enables zero-shot task generalization.
\newblock In \emph{International Conference on Learning Representations}, 2022.

\bibitem[Shridhar et~al.(2020)Shridhar, Thomason, Gordon, Bisk, Han, Mottaghi,
  Zettlemoyer, and Fox]{ALFRED20}
Mohit Shridhar, Jesse Thomason, Daniel Gordon, Yonatan Bisk, Winson Han,
  Roozbeh Mottaghi, Luke Zettlemoyer, and Dieter Fox.
\newblock {ALFRED: A Benchmark for Interpreting Grounded Instructions for
  Everyday Tasks}.
\newblock In \emph{The IEEE Conference on Computer Vision and Pattern
  Recognition (CVPR)}, 2020.
\newblock URL \url{https://arxiv.org/abs/1912.01734}.

\bibitem[Sinha et~al.(2022)Sinha, Mandlekar, and Garg]{sinha2022s4rl}
Samarth Sinha, Ajay Mandlekar, and Animesh Garg.
\newblock S4rl: Surprisingly simple self-supervision for offline reinforcement
  learning in robotics.
\newblock In \emph{Conference on Robot Learning}, pages 907--917. PMLR, 2022.

\bibitem[Sodhani et~al.(2021)Sodhani, Zhang, and Pineau]{care}
Shagun Sodhani, Amy Zhang, and Joelle Pineau.
\newblock Multi-task reinforcement learning with context-based representations.
\newblock In \emph{International Conference on Machine Learning}, pages
  9767--9779. PMLR, 2021.

\bibitem[Sohl-Dickstein et~al.(2015)Sohl-Dickstein, Weiss, Maheswaranathan, and
  Ganguli]{thermodynamic}
Jascha Sohl-Dickstein, Eric Weiss, Niru Maheswaranathan, and Surya Ganguli.
\newblock Deep unsupervised learning using nonequilibrium thermodynamics.
\newblock In Francis Bach and David Blei, editors, \emph{Proceedings of the
  32nd International Conference on Machine Learning}, volume~37 of
  \emph{Proceedings of Machine Learning Research}, pages 2256--2265, Lille,
  France, 07--09 Jul 2015. PMLR.
\newblock URL \url{https://proceedings.mlr.press/v37/sohl-dickstein15.html}.

\bibitem[Song et~al.(2023)Song, Dhariwal, Chen, and
  Sutskever]{song2023consistency}
Yang Song, Prafulla Dhariwal, Mark Chen, and Ilya Sutskever.
\newblock Consistency models, 2023.

\bibitem[Sun et~al.(2022)Sun, Zhang, Xu, and Tomizuka]{sun2022paco}
Lingfeng Sun, Haichao Zhang, Wei Xu, and Masayoshi Tomizuka.
\newblock Paco: Parameter-compositional multi-task reinforcement learning.
\newblock In Alice~H. Oh, Alekh Agarwal, Danielle Belgrave, and Kyunghyun Cho,
  editors, \emph{Advances in Neural Information Processing Systems}, 2022.

\bibitem[Sun et~al.(2023)Sun, Ma, Madaan, Bonatti, Huang, and
  Kapoor]{sun2023smart}
Yanchao Sun, Shuang Ma, Ratnesh Madaan, Rogerio Bonatti, Furong Huang, and
  Ashish Kapoor.
\newblock {SMART}: Self-supervised multi-task pretraining with control
  transformers.
\newblock In \emph{The Eleventh International Conference on Learning
  Representations}, 2023.

\bibitem[Taiga et~al.(2023)Taiga, Agarwal, Farebrother, Courville, and
  Bellemare]{taiga2023investigating}
Adrien~Ali Taiga, Rishabh Agarwal, Jesse Farebrother, Aaron Courville, and
  Marc~G Bellemare.
\newblock Investigating multi-task pretraining and generalization in
  reinforcement learning.
\newblock In \emph{The Eleventh International Conference on Learning
  Representations}, 2023.
\newblock URL \url{https://openreview.net/forum?id=sSt9fROSZRO}.

\bibitem[Tarasov et~al.(2022)Tarasov, Nikulin, Akimov, Kurenkov, and
  Kolesnikov]{tarasov2022corl}
Denis Tarasov, Alexander Nikulin, Dmitry Akimov, Vladislav Kurenkov, and Sergey
  Kolesnikov.
\newblock {CORL}: Research-oriented deep offline reinforcement learning
  library.
\newblock In \emph{3rd Offline RL Workshop: Offline RL as a ''Launchpad''},
  2022.
\newblock URL \url{https://openreview.net/forum?id=SyAS49bBcv}.

\bibitem[Touvron et~al.(2023)Touvron, Lavril, Izacard, Martinet, Lachaux,
  Lacroix, Rozi{\`e}re, Goyal, Hambro, Azhar, et~al.]{touvron2023llama}
Hugo Touvron, Thibaut Lavril, Gautier Izacard, Xavier Martinet, Marie-Anne
  Lachaux, Timoth{\'e}e Lacroix, Baptiste Rozi{\`e}re, Naman Goyal, Eric
  Hambro, Faisal Azhar, et~al.
\newblock Llama: Open and efficient foundation language models.
\newblock \emph{arXiv preprint arXiv:2302.13971}, 2023.

\bibitem[Van~der Maaten and Hinton(2008)]{tsne}
Laurens Van~der Maaten and Geoffrey Hinton.
\newblock Visualizing data using t-sne.
\newblock \emph{Journal of machine learning research}, 9\penalty0 (11), 2008.

\bibitem[Vaswani et~al.(2017)Vaswani, Shazeer, Parmar, Uszkoreit, Jones, Gomez,
  Kaiser, and Polosukhin]{transformer}
Ashish Vaswani, Noam Shazeer, Niki Parmar, Jakob Uszkoreit, Llion Jones,
  Aidan~N Gomez, {\L}ukasz Kaiser, and Illia Polosukhin.
\newblock Attention is all you need.
\newblock \emph{Advances in neural information processing systems}, 30, 2017.

\bibitem[Wang et~al.(2023)Wang, Hunt, and Zhou]{diffusionql}
Zhendong Wang, Jonathan~J Hunt, and Mingyuan Zhou.
\newblock Diffusion policies as an expressive policy class for offline
  reinforcement learning.
\newblock In \emph{The Eleventh International Conference on Learning
  Representations}, 2023.

\bibitem[Wei et~al.(2022)Wei, Bosma, Zhao, Guu, Yu, Lester, Du, Dai, and
  Le]{wei2022finetuned}
Jason Wei, Maarten Bosma, Vincent Zhao, Kelvin Guu, Adams~Wei Yu, Brian Lester,
  Nan Du, Andrew~M. Dai, and Quoc~V Le.
\newblock Finetuned language models are zero-shot learners.
\newblock In \emph{International Conference on Learning Representations}, 2022.

\bibitem[Wen et~al.(2022)Wen, Wan, Zhou, Hou, Cao, Le, Chen, Tian, Zhang, and
  Wang]{wen2022realization}
Ying Wen, Ziyu Wan, Ming Zhou, Shufang Hou, Zhe Cao, Chenyang Le, Jingxiao
  Chen, Zheng Tian, Weinan Zhang, and Jun Wang.
\newblock On realization of intelligent decision-making in the real world: A
  foundation decision model perspective.
\newblock \emph{arXiv preprint arXiv:2212.12669}, 2022.

\bibitem[Wu et~al.(2023)Wu, Yin, Qi, Wang, Tang, and Duan]{visualGPT}
Chenfei Wu, Shengming Yin, Weizhen Qi, Xiaodong Wang, Zecheng Tang, and Nan
  Duan.
\newblock Visual chatgpt: Talking, drawing and editing with visual foundation
  models.
\newblock \emph{arXiv preprint arXiv:2303.04671}, 2023.

\bibitem[Xiao et~al.(2022)Xiao, Kreis, and Vahdat]{xiao2022tackling}
Zhisheng Xiao, Karsten Kreis, and Arash Vahdat.
\newblock Tackling the generative learning trilemma with denoising diffusion
  gans.
\newblock In \emph{International Conference on Learning Representations}, 2022.

\bibitem[Xu et~al.(2022{\natexlab{a}})Xu, Zhan, Yin, and Qin]{pmlr-v162-xu22l}
Haoran Xu, Xianyuan Zhan, Honglei Yin, and Huiling Qin.
\newblock Discriminator-weighted offline imitation learning from suboptimal
  demonstrations.
\newblock In \emph{Proceedings of the 39th International Conference on Machine
  Learning}, volume 162 of \emph{Proceedings of Machine Learning Research},
  pages 24725--24742. PMLR, 17--23 Jul 2022{\natexlab{a}}.

\bibitem[Xu et~al.(2022{\natexlab{b}})Xu, Shen, Zhang, Lu, Zhao, Tenenbaum, and
  Gan]{promptdt}
Mengdi Xu, Yikang Shen, Shun Zhang, Yuchen Lu, Ding Zhao, Joshua Tenenbaum, and
  Chuang Gan.
\newblock Prompting decision transformer for few-shot policy generalization.
\newblock In Kamalika Chaudhuri, Stefanie Jegelka, Le~Song, Csaba Szepesvari,
  Gang Niu, and Sivan Sabato, editors, \emph{Proceedings of the 39th
  International Conference on Machine Learning}, volume 162 of
  \emph{Proceedings of Machine Learning Research}, pages 24631--24645. PMLR,
  17--23 Jul 2022{\natexlab{b}}.
\newblock URL \url{https://proceedings.mlr.press/v162/xu22g.html}.

\bibitem[Yang et~al.(2020)Yang, Xu, Wu, and Wang]{softmodular}
Ruihan Yang, Huazhe Xu, Yi~Wu, and Xiaolong Wang.
\newblock Multi-task reinforcement learning with soft modularization.
\newblock \emph{Advances in Neural Information Processing Systems},
  33:\penalty0 4767--4777, 2020.

\bibitem[Yarats et~al.(2021)Yarats, Kostrikov, and Fergus]{drq}
Denis Yarats, Ilya Kostrikov, and Rob Fergus.
\newblock Image augmentation is all you need: Regularizing deep reinforcement
  learning from pixels.
\newblock In \emph{International Conference on Learning Representations}, 2021.
\newblock URL \url{https://openreview.net/forum?id=GY6-6sTvGaf}.

\bibitem[Yu and Mooney(2023)]{yu2023using}
Albert Yu and Ray Mooney.
\newblock Using both demonstrations and language instructions to efficiently
  learn robotic tasks.
\newblock In \emph{The Eleventh International Conference on Learning
  Representations}, 2023.
\newblock URL \url{https://openreview.net/forum?id=4u42KCQxCn8}.

\bibitem[Yu et~al.(2019)Yu, Quillen, He, Julian, Hausman, Finn, and
  Levine]{metaworld}
Tianhe Yu, Deirdre Quillen, Zhanpeng He, Ryan Julian, Karol Hausman, Chelsea
  Finn, and Sergey Levine.
\newblock Meta-world: A benchmark and evaluation for multi-task and meta
  reinforcement learning.
\newblock In \emph{Conference on Robot Learning (CoRL)}, 2019.
\newblock URL \url{https://arxiv.org/abs/1910.10897}.

\bibitem[Yu et~al.(2020)Yu, Kumar, Gupta, Levine, Hausman, and Finn]{PCG_grad}
Tianhe Yu, Saurabh Kumar, Abhishek Gupta, Sergey Levine, Karol Hausman, and
  Chelsea Finn.
\newblock Gradient surgery for multi-task learning.
\newblock In H.~Larochelle, M.~Ranzato, R.~Hadsell, M.F. Balcan, and H.~Lin,
  editors, \emph{Advances in Neural Information Processing Systems}, volume~33,
  pages 5824--5836. Curran Associates, Inc., 2020.
\newblock URL
  \url{https://proceedings.neurips.cc/paper_files/paper/2020/file/3fe78a8acf5fda99de95303940a2420c-Paper.pdf}.

\bibitem[Yu et~al.(2021)Yu, Kumar, Chebotar, Hausman, Levine, and
  Finn]{data-share}
Tianhe Yu, Aviral Kumar, Yevgen Chebotar, Karol Hausman, Sergey Levine, and
  Chelsea Finn.
\newblock Conservative data sharing for multi-task offline reinforcement
  learning.
\newblock \emph{Advances in Neural Information Processing Systems},
  34:\penalty0 11501--11516, 2021.

\bibitem[Yu et~al.(2023)Yu, Xiao, Stone, Tompson, Brohan, Wang, Singh, Tan,
  Peralta, Ichter, et~al.]{yu2023scaling}
Tianhe Yu, Ted Xiao, Austin Stone, Jonathan Tompson, Anthony Brohan, Su~Wang,
  Jaspiar Singh, Clayton Tan, Jodilyn Peralta, Brian Ichter, et~al.
\newblock Scaling robot learning with semantically imagined experience.
\newblock \emph{arXiv preprint arXiv:2302.11550}, 2023.

\bibitem[Yuan and Lu(2022)]{context-1}
Haoqi Yuan and Zongqing Lu.
\newblock Robust task representations for offline meta-reinforcement learning
  via contrastive learning.
\newblock In \emph{International Conference on Machine Learning}, pages
  25747--25759. PMLR, 2022.

\bibitem[Zhang et~al.(2022)Zhang, Xu, and Yu]{zhang2022generative}
Haichao Zhang, Wei Xu, and Haonan Yu.
\newblock Generative planning for temporally coordinated exploration in
  reinforcement learning.
\newblock In \emph{International Conference on Learning Representations}, 2022.
\newblock URL \url{https://openreview.net/forum?id=YZHES8wIdE}.

\bibitem[Zintgraf et~al.(2020)Zintgraf, Shiarlis, Igl, Schulze, Gal, Hofmann,
  and Whiteson]{context-2}
Luisa Zintgraf, Kyriacos Shiarlis, Maximilian Igl, Sebastian Schulze, Yarin
  Gal, Katja Hofmann, and Shimon Whiteson.
\newblock Varibad: A very good method for bayes-adaptive deep rl via
  meta-learning.
\newblock In \emph{International Conference on Learning Representations}, 2020.

\end{thebibliography}

\newpage
\appendix

\section{The Details of \textsc{MTDiff}}
\label{appendix:mtdiff}
In this section, we give the pseudocodes of \textsc{MTDiff-p} and \textsc{MTDiff-s} in Alg.~\ref{alg:diff-p} and Alg.~\ref{alg:diff-s}, respectively. Then we describe various details of the training process, architecture and hyperparameters:
\begin{itemize}[leftmargin=*]
    \item We set the per-task batch size as 8, so the total batch size is 400. We train our model using Adam optimizer \cite{kingma2014adam} with $2e^{-4}$ learning rate for $2e^6$ train steps.
    \item We train \textsc{MTDiff} on NVIDIA GeForce RTX 3080 for around 50 hours.
    \item We represent the noise model as the transformer-based architecture described in Section 3.3. MLP $f_P$ which processes prompt is a 3-layered MLP (prepended by a layer norm \cite{ba2016layer} and with Mish activation). MLP $f_{Ti}$ which processes diffusion timestep $f_R$ is a 2-layered MLP (prepended by a Sinusoidal embedding and with Mish activation). $f_R$ which processes conditioned Return and $f_H$ which processes state history are 3-layered MLPs with Mish activation. $f_A$ which processes actions, $f_{TR}$ which process transitions and prediction head are 2-layered MLPs with Mish activation. The GPT2 transformer is configured as 6 hidden layers and 4 attention heads. The code of GPT2 is borrowed from \url{https://github.com/kzl/decision-transformer}.
    \item We choose the probability $p$ of removing the conditioning information to be 0.25.
    \item In \textsc{MTDiff-p}, we choose the state history length $L=2$ for Meta-World and $L=5$ for Maze2D.
    \item We choose the trajectory prompt length $J=20$.
    \item We use $K=200$ for diffusion steps.
    \item We set guidance scale $\alpha=1.2$ for extracting near-optimal behavior.
    \item We choose $\beta=0.5$ for low temperature sampling.
\end{itemize}
\begin{algorithm}[!htbp]
    \caption{\textsc{MTDiff-p} Training and Evaluation}
    \label{alg:diff-p}
    \emph{\# Training Process} \\
    \textbf{Initialize}: training tasks $\mathcal{T}^{train}$, training iterations $N$, multi-task dataset $\mathcal{D}$, per-task batch size $M$, multi-task trajectory prompts $Z$
    \begin{algorithmic}[1]
    \FOR{$n=1$ \TO $N$}
        \FOR{Each task $\mathcal{T}_i\in \mathcal{T}^{train}$}
             \STATE Sample action sequences $\bm{x}^{\textcolor{red}{p}}_0(\tau_i)$ of length $H$ and corresponding state history $S^{\rm prev}_i$ of length $L$ from $\mathcal{D}_i$ with batch size $M$
             \STATE Compute normalized return $R(\tau_i)$ under $\tau_i$
             \STATE Sample trajectory prompts $\tau^*_i$ of length $J$ from $Z_i$ with batch size $M$
        \ENDFOR
        \STATE Get a batch $\mathcal{B}=\{\bm{x}^{\textcolor{red}{p}}_0(\tau_i),\tau^*_i,S^{\rm prev}_i,R(\tau_i)\}_{i=1}^{|\mathcal{T}^{train}|}$
        \STATE Randomly sample a diffusion timestep $k\sim \mathcal{U}(1, K)$ and obtain noisy sequences $\bm{x}^{\textcolor{red}{p}}_k(\tau_i)$
        \STATE Omit the $R(\tau)$ condition with probability $\beta\sim {\rm{Bern}}(p)$
        \STATE Compute $ \mathcal{L}^{\textcolor{red}{p}}(\theta)$ and update \textsc{MTDiff-p} model
    \ENDFOR
    \end{algorithmic}
    \emph{\# Evaluation Process} \\ 
    \vspace{-1.2em}
    \begin{algorithmic}[1] 
    \STATE Given a task, reset the environment and set desired return $R_{\rm max}(\tau)$
    \STATE Obtain the initial state history $h_0$, few-shot prompts $Z$
    \STATE Set low-temperature sampling scale $\beta$, classifier-free guidance scale $\alpha$
    \FOR{$t =0$ \TO $t_{\rm max}$}
    \STATE Initialize $\bm{x}^{\textcolor{red}{p}}_K(\tau)\sim\mathcal{N(\boldsymbol{\mathrm{0}},\beta\boldsymbol{{I}})}$
    \STATE Sample $\tau^*\sim Z$, and formulate $\bm{y}'(\tau) = [h_t, \tau^*]$
    \FOR{$k=K$ \TO $1$}
    \STATE $\hat{\epsilon} = \epsilon_\theta\big(\bm{x}^{\textcolor{red}{p}}_k(\tau),\bm{y}'(\tau), \varnothing, k\big)+\alpha\big(\epsilon_\theta(\bm{x}^{\textcolor{red}{p}}_k(\tau), \bm{y}'(\tau), R_{\rm max}(\tau), k)-\epsilon_\theta(\bm{x}^{\textcolor{red}{p}}_k(\tau), \bm{y}'(\tau), \varnothing, k)\big)$
    \STATE $(\mu_{k-1},\Sigma_{k-1})\gets {\rm{Denoise}}(\bm{x}^{\textcolor{red}{p}}_k(\tau),\hat{\epsilon})$
    \STATE $\bm{x}^{\textcolor{red}{p}}_{k-1}(\tau)\sim\mathcal{N}(\mu_{k-1},\beta\Sigma_{k-1})$
    \ENDFOR
    \STATE Execute the first action from $\bm{x}^{\textcolor{red}{p}}_0(\tau)$ as the current action to interact with the environment
    \STATE Obtain the next state, and update $h_t$
    \ENDFOR
    \end{algorithmic}
\end{algorithm}
\begin{algorithm}[htbp]
    \caption{\textsc{MTDiff-s} Training and Data Synthesis}
    \label{alg:diff-s}
    \emph{\# Training Process} \\
    \textbf{Initialize}: training tasks $\mathcal{T}^{train}$, training iterations $N$, multi-task dataset $\mathcal{D}$, per-task batch size $M$, multi-task trajectory prompts $Z$
    \begin{algorithmic}[1]
    \FOR{$n=1$ \TO $N$}
        \FOR{Each task $\mathcal{T}_i\in \mathcal{T}^{train}$}
             \STATE Sample transition sequences $\bm{x}^{\textcolor{red}{s}}_0(\tau_i)$ of length $H$ from $\mathcal{D}_i$ with batch size $M$
             \STATE Sample trajectory prompts $\tau^*_i$ of length $J$ from $Z_i$ with batch size $M$
        \ENDFOR
        \STATE Get a batch $\mathcal{B}=\{\bm{x}^{\textcolor{red}{s}}_0(\tau_i),\tau^*_i\}_{i=1}^{|\mathcal{T}^{train}|}$
        \STATE Randomly sample a diffusion timestep $k\sim \mathcal{U}(1, K)$ and obtain noisy sequences $\bm{x}^{\textcolor{red}{s}}_k(\tau_i)$
        \STATE Compute $ \mathcal{L}^{\textcolor{red}{s}}(\theta)$ and update \textsc{MTDiff-s} model
    \ENDFOR
    \end{algorithmic}
    \emph{\# Data Synthesis Process} \\ 
    \textbf{Initialize}: synthetic dataset $\mathcal{D}=\varnothing$, synthesizing times $M$
    \begin{algorithmic}[1] 
    \STATE Given a task, obtain few-shot prompts $Z$
    \FOR{$m =1$ \TO $M$}
    \STATE Initialize $\bm{x}^{\textcolor{red}{s}}_K(\tau)\sim\mathcal{N(\boldsymbol{\mathrm{0}},\boldsymbol{{I}})}$
    \STATE Sample $\tau^*\sim Z$, and formulate $\bm{y}^{\textcolor{red}{s}}_k(\tau) = [\tau^*]$
    \FOR{$k=K$ \TO $1$}
    \STATE $\hat{\epsilon} = \epsilon_\theta\big(\bm{x}^{\textcolor{red}{s}}_k(\tau),\bm{y}^{\textcolor{red}{s}}(\tau), k\big)$
    \STATE $(\mu_{k-1},\Sigma_{k-1})\gets {\rm{Denoise}}(\bm{x}^{\textcolor{red}{s}}_k(\tau),\hat{\epsilon})$
    \STATE $\bm{x}^{\textcolor{red}{s}}_{k-1}(\tau)\sim\mathcal{N}(\mu_{k-1},\Sigma_{k-1})$
    \ENDFOR
    \STATE Update $\mathcal{D}=\mathcal{D}\cup\bm{x}^{\textcolor{red}{s}}_0(\tau)$
    \ENDFOR
    \end{algorithmic}
\end{algorithm}
\section{The Details of Baselines}
\label{appendix:algo}
In this section, we describe the implementation details of the baselines:
\begin{itemize}[leftmargin=*]
    \item \textbf{PromptDT} uses the same prompts and GPT2 transformer in \textsc{MTDiff-p} for taining. We borrow the code from \url{https://github.com/mxu34/prompt-dt} for implementation.
    \item \textbf{MTDT} embeds the tskaID which indicates the task to a embedding $z$ with size 12, then $z$ is concatenated with the raw state. With such conditioning, we broaden the original state space to equip DT with the ability to identify tasks in this multi-task setting. We keep other hyperparameters and implementation details the same as the official version \url{https://github.com/kzl/decision-transformer/}.
    \item \textbf{MTIQL} uses a multi-head critic network to predict the $Q$ value for each task, and each head is parameterized with a 3-layered MLP (with Mish activation). The actor-network is parameterized with a 3-layered MLP (with Mish activation). During training and inference, the scalar taskID is embedded via 3-layered MLP (with Mish activation) into latent variable $z$, and the input of the actor becomes the concatenation of the original state and $z$. We build MTIQL based on the code \url{https://github.com/tinkoff-ai/CORL} \cite{tarasov2022corl}.
    \item \textbf{MTCQL} is applied with a similar revision in MTIQL. The main difference is that MTCQL is based on CQL \cite{cql} algorithm instead of the IQL algorithm \cite{iql}. We build MTCQL based on the code \url{https://github.com/tinkoff-ai/CORL} \cite{tarasov2022corl}.
    \item \textbf{MTBC} uses a similar taskID-cognitional actor in MTIQL and MTCQL. For training and inference, the scalar taskID is embedded via a 3-layered MLP (with Mish activation) into latent variable $z$, and the input of the actor becomes the concatenation of the original state and $z$. The actor is parameterized with a 3-layered MLP and outputs predicted actions.
    \item \textbf{RAD} adopts the random amplitude scaling \cite{laskin2020reinforcement} that multiplies a random variable to states during training, i.e., $s' = s \times z$, where $z \sim {\rm Uniform}[\alpha,\beta]$. We choose $\alpha=0.8$ and $\beta=1.2$.
    \item \textbf{S4RL} adopts the adversarial state training \cite{sinha2022s4rl} by taking gradients with respect to the value function to obtain a new state, i.e. $s' \gets s + \epsilon\nabla_s\mathbb{J}_Q(\pi(s))$, where $\mathbb{J}_Q$ is the policy evaluation update performed via a $Q$ function, and $\epsilon$ is the size of gradient steps. We choose $\epsilon=0.01$.
\end{itemize}
\section{Ablation Study on Model Architecture}
\label{appendix:arch}
The architecture described in \S3.3 handles input types of different modalities as tokens that share similar formats, actively capturing interactions between modalities. The incorporation of a transformer is also helpful for sequential modeling. To ablate the effectiveness of our architecture design, we train \textsc{MTDiff-p} using U-Net with a similar model size to ours on the near-optimal dataset. We use a Temporal Convolutional Network (TCN) \cite{tcn} to encode the prompt into an embedding $z_p$, and inject it in the U-Net layers as a condition. We follow the conditional approach in \cite{decisiondiffuser} and borrow the code for temporal U-Net from \url{https://github.com/jannerm/diffuser} \cite{diffuser}. The results summarized in Table \ref{table:ablation} show that our model architecture outperforms U-Net to learn from multi-task datasets.
 \begin{table}[!h]
  \begin{center}
    \caption{Average success rate across 3 different seeds of \textsc{MTDiff-p} and \textsc{MTDiff-p} (U-Net) on MT50-rand.}
    \resizebox{\linewidth}{!}{
    \label{table:ablation}
    \begin{tabular}{c|cc}
    \hline
   \textbf{Methods}& \textbf{Success rate on near-optimal dataset (\%)} &\textbf{Success rate on sub-optimal dataset (\%) }\\
    \hline
      \textsc{MTDiff-p} & \bm{$59.53\pm1.12$}&\bm{$48.67\pm1.32$}\\
      \textsc{MTDiff-p} (U-Net)& $55.67\pm1.27$&$47.42\pm0.74$\\
      \hline
    \end{tabular}
    }
  \end{center}
\end{table}
\section{Environmental Details of Maze2D}
\label{appendix:maze2d}
\begin{figure}[htbp]
\centering
    \subfigure[map 1]{
        \includegraphics[width=0.24\linewidth]{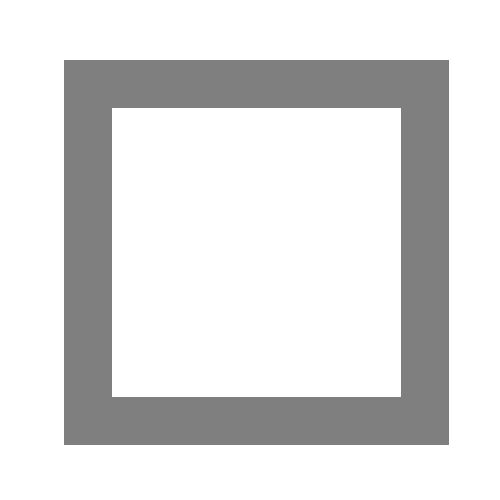}
    }\hspace{-2.3mm}
    \subfigure[map 2]{
	\includegraphics[width=0.24\linewidth]{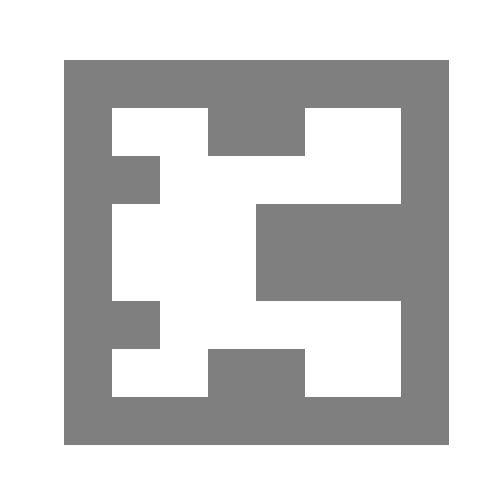}
    }\hspace{-2.3mm}
    \subfigure[map 3]{
	\includegraphics[width=0.24\linewidth]{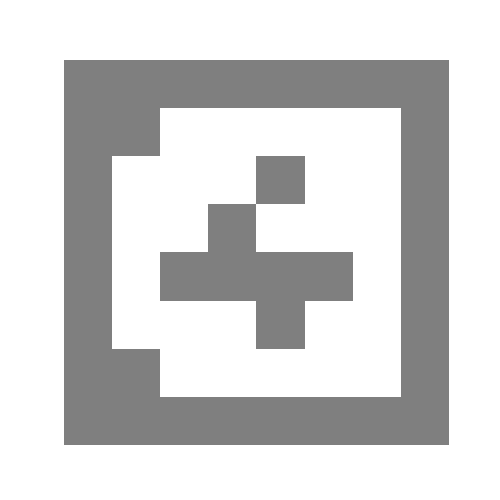}
    }
    \hspace{-2.3mm}
    \subfigure[map 4]{
	\includegraphics[width=0.24\linewidth]{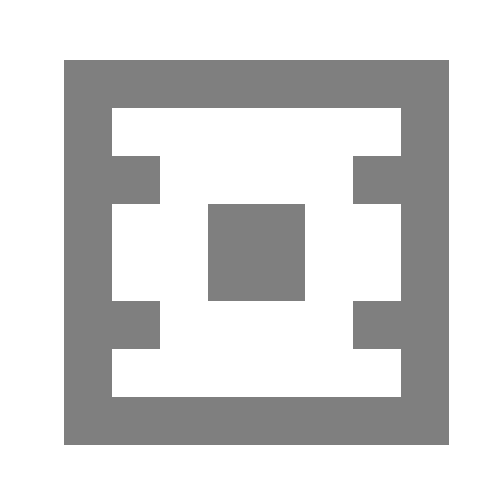}
    }
    \\
    \subfigure[map 5]{
        \includegraphics[width=0.24\linewidth]{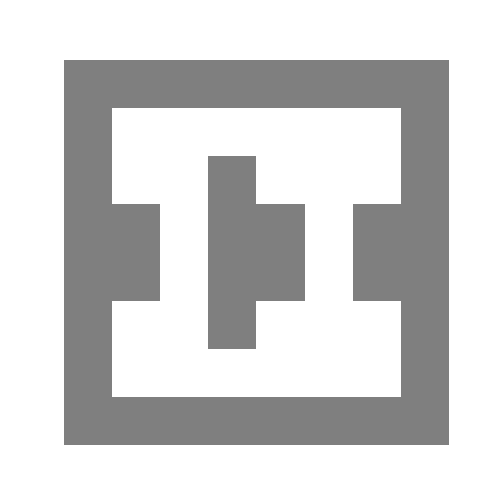}
    }\hspace{-2.3mm}
    \subfigure[map 6]{
	\includegraphics[width=0.24\linewidth]{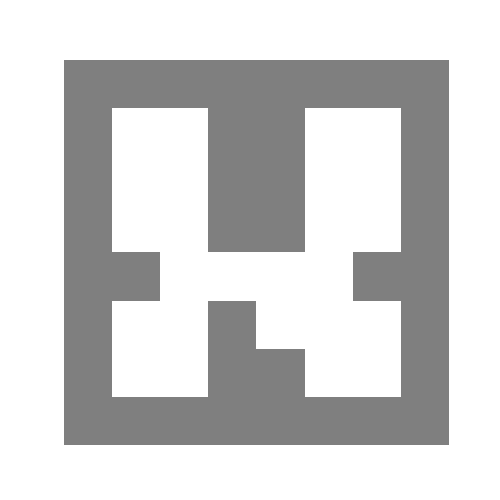}
    }\hspace{-2.3mm}
    \subfigure[map 7]{
	\includegraphics[width=0.24\linewidth]{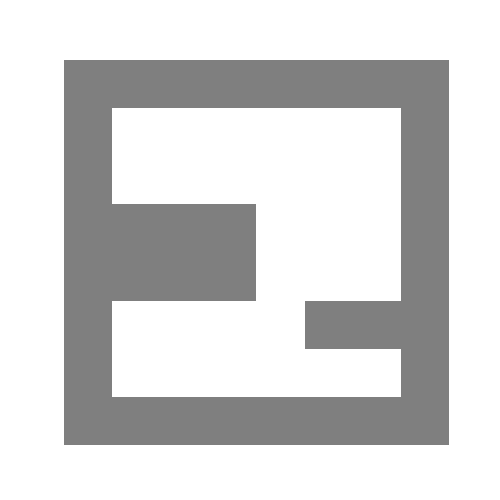}
    }
    \hspace{-2.3mm}
    \subfigure[map 8]{
	\includegraphics[width=0.24\linewidth]{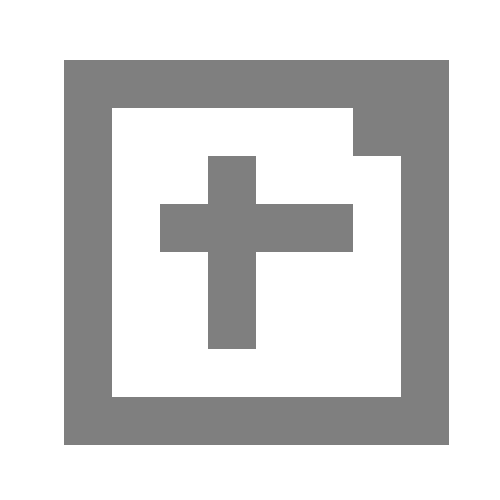}
    }
    \caption{2D visualization of eight training maps designed in Maze2D. }
    \label{fig:train_map}
\end{figure}
\begin{figure}[htbp]
\centering
    \subfigure[PromptDT]{
        \includegraphics[width=0.24\linewidth]{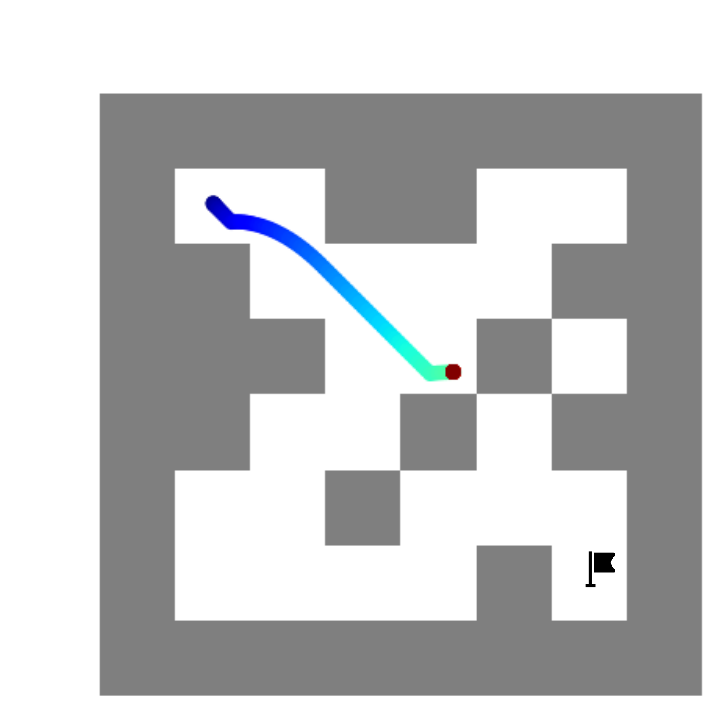}
    }\hspace{-2.3mm}
    \subfigure[\textsc{MTDiff}]{
	\includegraphics[width=0.24\linewidth]{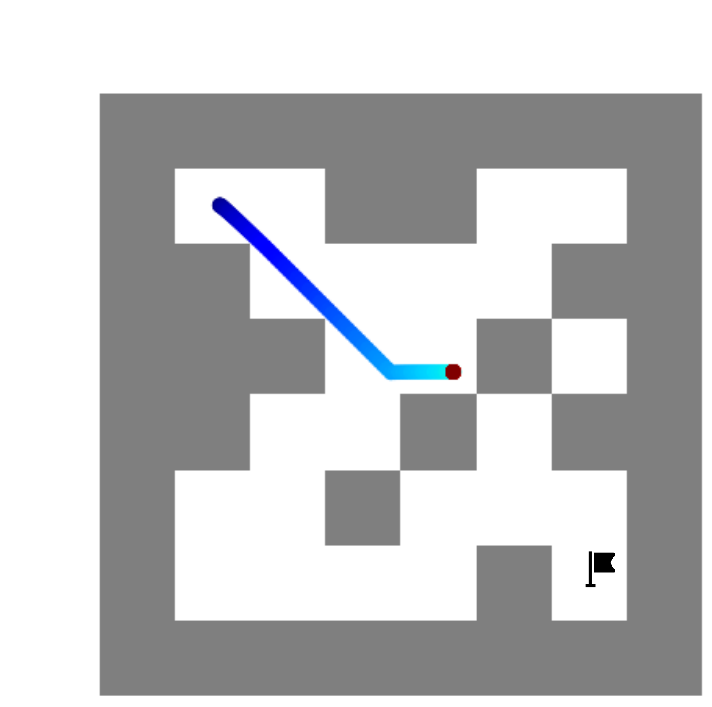}
    }\hspace{-2.3mm}
    \subfigure[PromptDT]{
	\includegraphics[width=0.24\linewidth]{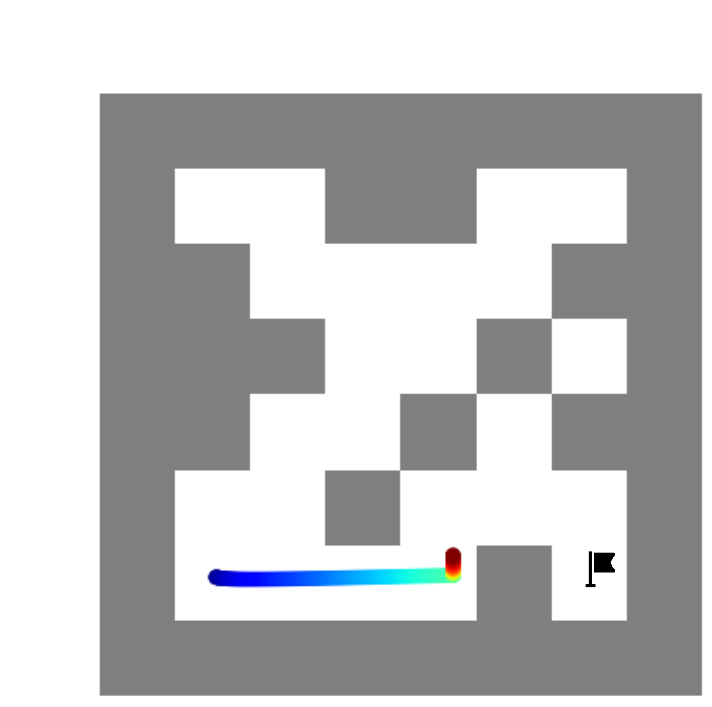}
    }
    \hspace{-2.3mm}
    \subfigure[\textsc{MTDiff}]{
	\includegraphics[width=0.24\linewidth]{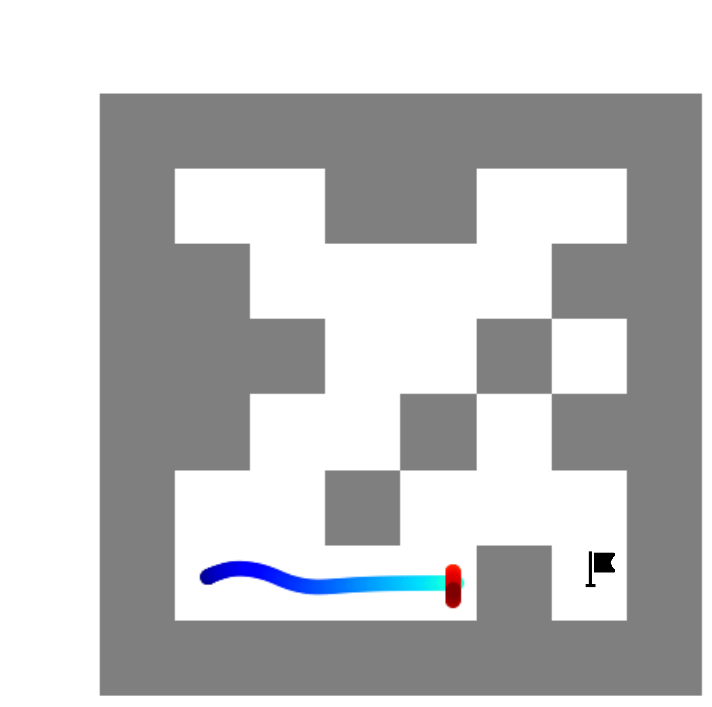}
    }
   \ \ \ \ \ \ \ \ \ \ \ \ \ \ \ \ \ \ \ \ \ \ \textbf{(1) Failure case 1} \ \ \ \ \ \ \ \ \ \ \ \ \ \ \ \ \ \ \ \ \ \ \ \ \ \ \ \ \ \ \ \ \ \ \ \ \ \ \ \ \ \ \ \ \ \ \ \textbf{(2) Failure case 2}
    \caption{2D visualization of 2 difficult cases where both PromptDT and \textsc{MTDiff} both fail. These 2 cases are both unseen during training. Goal position is denoted as {\includegraphics[width=0.015\linewidth]{images/flag.pdf}}.}
    \label{fig:failure}
\end{figure}
\begin{figure}[htbp]
\centering
    \subfigure[\textsc{MTDiff-p}]{
        \includegraphics[width=0.3\linewidth]{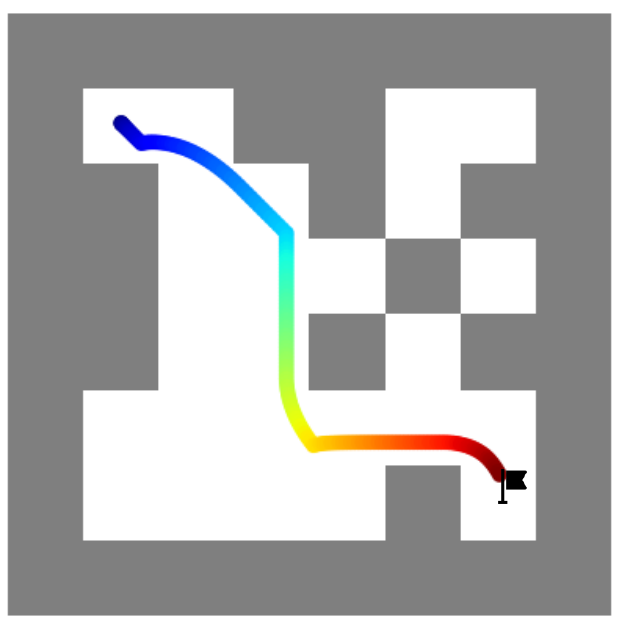}
    }\hspace{-2.3mm}
    \subfigure[PromptDT (failed)]{
	\includegraphics[width=0.3\linewidth]{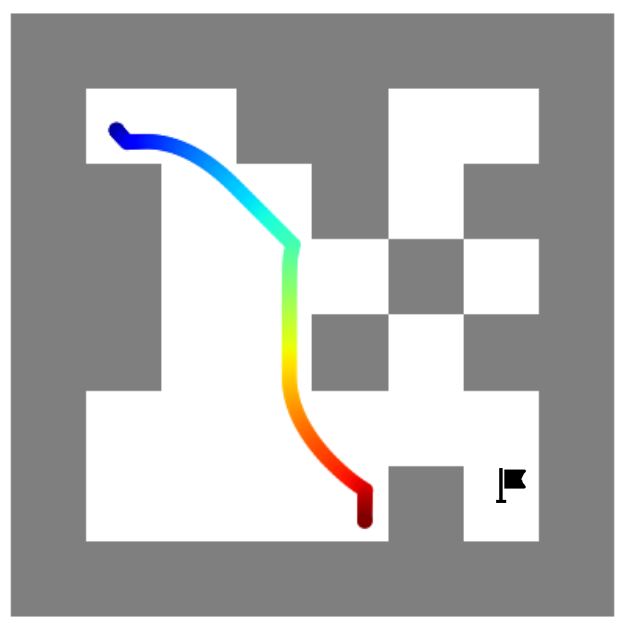}
    }\hspace{-2.3mm}
    \caption{Unseen maps of Maze2D with long planning path. \textsc{MTDiff-p} reach the goal while PromptDT fails. Goal position is denoted as {\includegraphics[width=0.015\linewidth]{images/flag.pdf}}.}
    \label{fig:another}
\end{figure}

We design eight different training maps for multi-task training, which are shown at Fig.~\ref{fig:train_map}. Different tasks have different reward functions and transition functions.
For generalizability evaluation, we have designed one new unseen map. Although in Fig.~4, \textsc{MTDiff} is able to generalize on new maps while PromptDT fails, we should acknowledged that \textsc{MTDiff} may fail at some difficult unseen cases, as shown in Fig.~\ref{fig:failure}. The reasons may lie in 2 folds: One is the inherent difficulty of the case itself, and the other is that the case's deviation from the distribution of the training cases surpasses the upper threshold of generalizability of \textsc{MTDiff}. We also provide another map where \textsc{MTDiff} succeeds while PromptDT fails in Fig.~\ref{fig:another}. 

We collect 35k episodes together to train our model and PromptDT. The episodic length is set as 600 for training and 200 for evaluation. After training with 512 batch size for 2e5 gradient steps, we evaluate these methods on seen and unseen maps.
\section{Data Analysis}
\label{appendix:vis}
\subsection{Distribution Visualization}
\begin{figure}[!htbp]
\centering
    \subfigure[Coffee-push]{
        \includegraphics[width=0.24\linewidth]{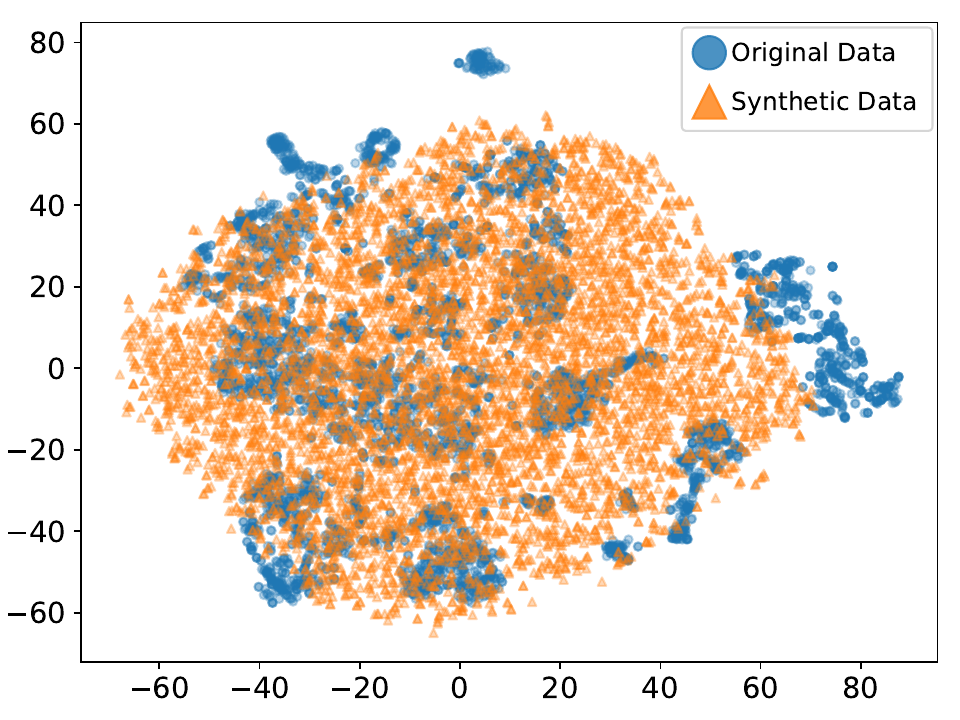}
    }\hspace{-2.3mm}
    \subfigure[Disassemble]{
	\includegraphics[width=0.24\linewidth]{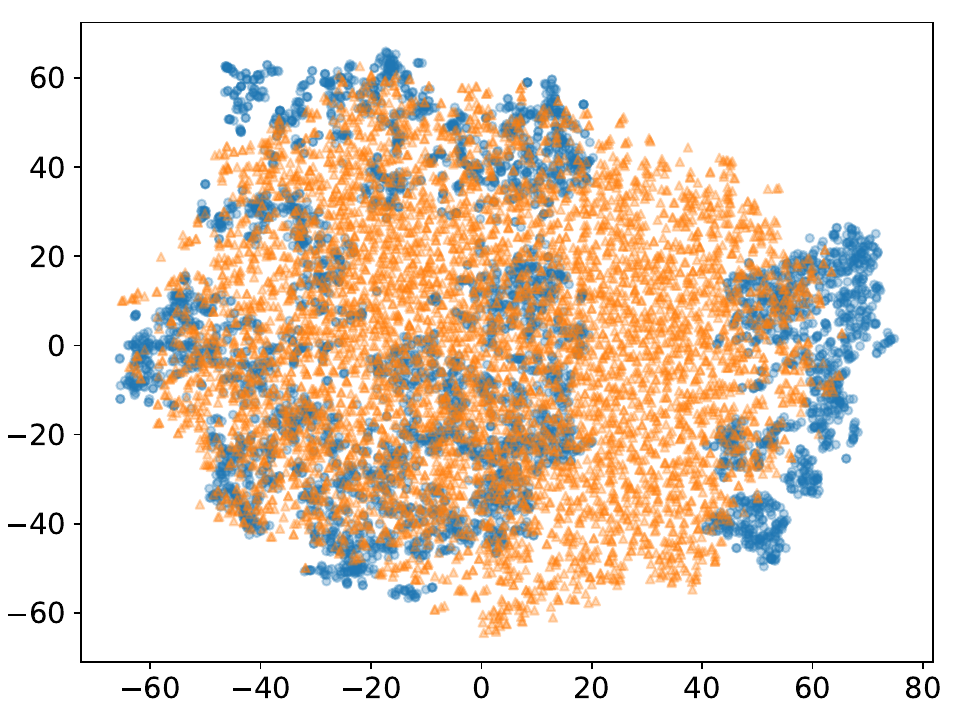}
    }\hspace{-2.3mm}
    \subfigure[hand-insert]{
	\includegraphics[width=0.24\linewidth]{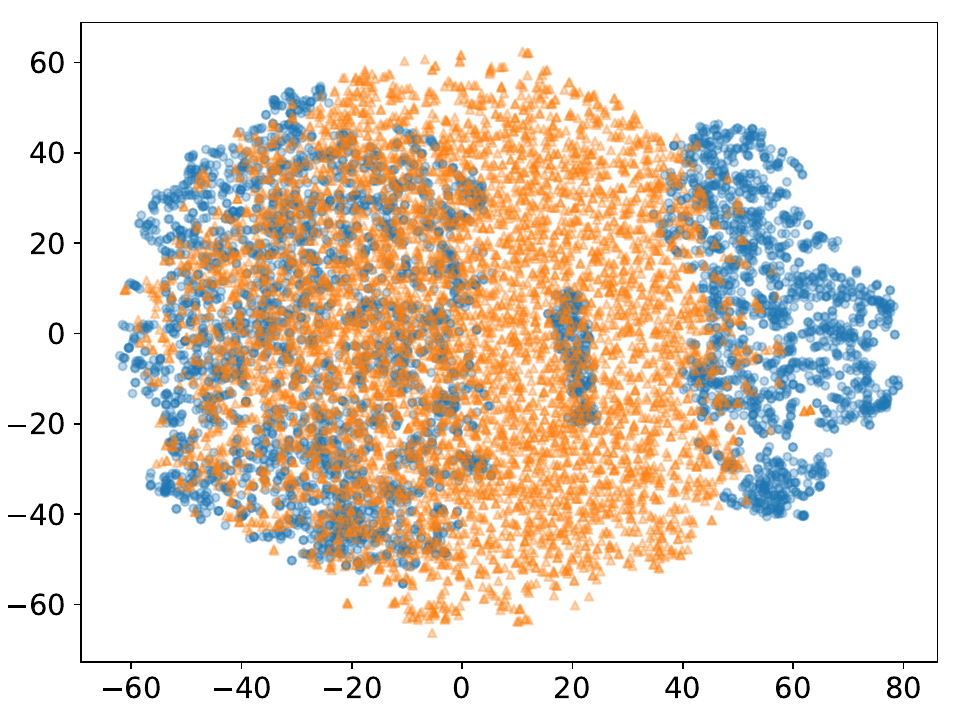}
    }
    \hspace{-2.3mm}
    \subfigure[box-close]{
	\includegraphics[width=0.24\linewidth]{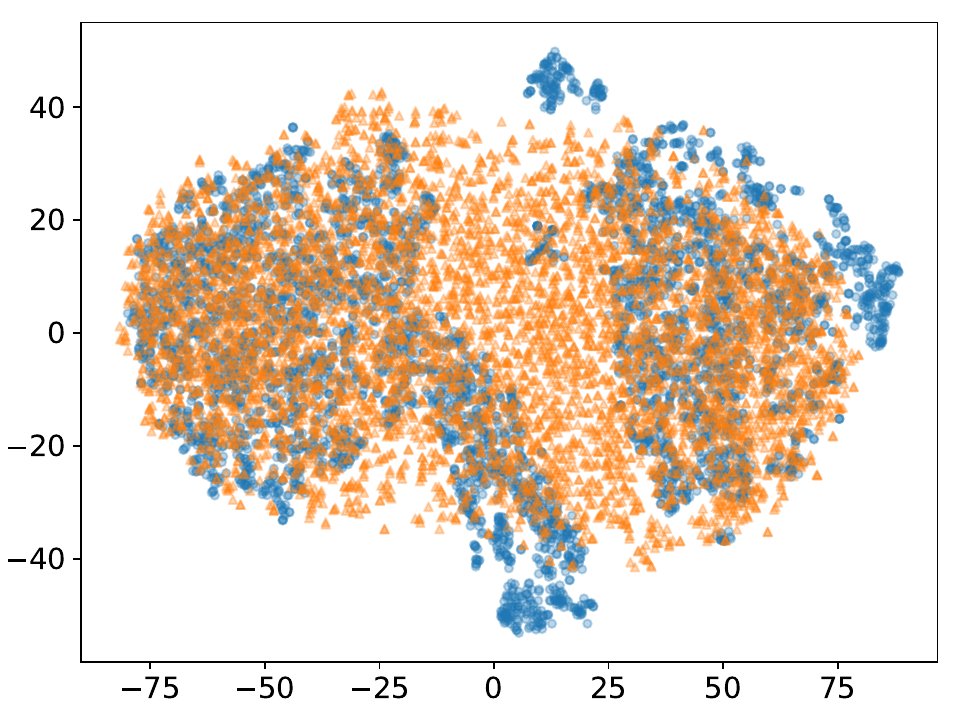}
    }
    \caption{2D visualization of sampled synthetic data and original data via T-SNE \cite{tsne}. The data is sampled from tasks coffee-push, disassemble, hand-insert and box-close respectively.}
    \label{fig:visual}
\end{figure}

 We find the synthetic data is high-fidelity, covering or even broadening the original data distribution, which makes the offline RL method performs better in the augmented dataset. The distribution visualization is shown in Fig.~\ref{fig:visual}.
\subsection{Statistical Analysis}
Following SynthER \cite{lu2023synthetic}, we measure the Dynamics Error (MSE between the augmented next state and true next state), and L2 Distance from Dataset (minimum L2 distance of each datapoint from the dataset) for the augmented data of each method (i.e., \textsc{MTDiff-s}, S4RL and RAD), as shown in Table~\ref{table:error}. Since S4RL performs data augmentation by adding data within the $\epsilon$-ball of the original data points, it has the smallest dynamics error with a small $\epsilon$. RAD performs random amplitude scaling and causes the largest dynamics error. We remark that S4RL performs local data augmentation around the original data and can be limited in expanding the data coverage of offline datasets. In contrast, our method generates data via diffusion model without explicit constraints to the original data points, which also has small dynamics error and significantly improves the data coverage, benefiting the offline RL training. 
\begin{table}[!ht]
    \centering
    \caption{Comparing L2 distance from the training dataset and dynamics error under each method.}
    \vspace{1em}
    \label{table:error}
    \begin{tabular}{c|c|c}
    \hline
        \textbf{Methods} & \textbf{Dynamics Error} & \textbf{L2 Distance from Dataset} \\ \hline
        \textsc{MTDiff-s} & $0.0174$ & $0.4552$ \\ \hline
        S4RL & $0.0001$ & $0.4024$ \\ \hline
        RAD & $0.0641$ & $0.4617$ \\ \hline
    \end{tabular}
\end{table}
\section{Limitations and Discussions}
\label{appendix:limi}
In this section, we will discuss the limitations and broader impacts of our proposed method \textsc{MTDiff}.
\paragraph{Limitation.}
Diffusion models are bottlenecked by their slow sampling speed, which caps the potential of \textsc{MTDiff} for real-time control. How to trade off the sampling speed and generative quality remains to be a crucial research topic. For a concrete example in MetaWorld, it takes on average 1.9s in wall-clock time to generate one action sequence for planning (hardware being a 3090 GPU). We can improve the inference speed by leveraging a recent sampler called DPM-solver \cite{lu2022dpmsolver,lu2022dpm++} to decrease the diffusion steps required to $0.2 \times$
 without any loss in performance, and using a larger batch size (leveraging the parallel computing power of GPUs) to evaluate multiple environments at once. Thus the evaluation run-time roughly matches the run-time of non-diffusion algorithms (diffusion step is 1). In addition, consistency models \cite{song2023consistency} are recently proposed to support one-step
and few-step generation, while the upper performance of what such models can achieve is still vague.
\paragraph{Broader Impacts.}
As far as we know, \textsc{MTDiff} is the first proposed diffusion-based approach for multi-task reinforcement learning. It could be applied to multi-task decision-making, and also could be used to synthesize more data to boost policy improvement. \textsc{MTDiff} provides a solution for achieving generalization in reinforcement learning.
\section{Dataset collection}
\label{appendix:dataset}
\begin{figure}[htbp]
\centering
\subfigure[Near-optimal dataset]{
        \includegraphics[width=0.49\linewidth]{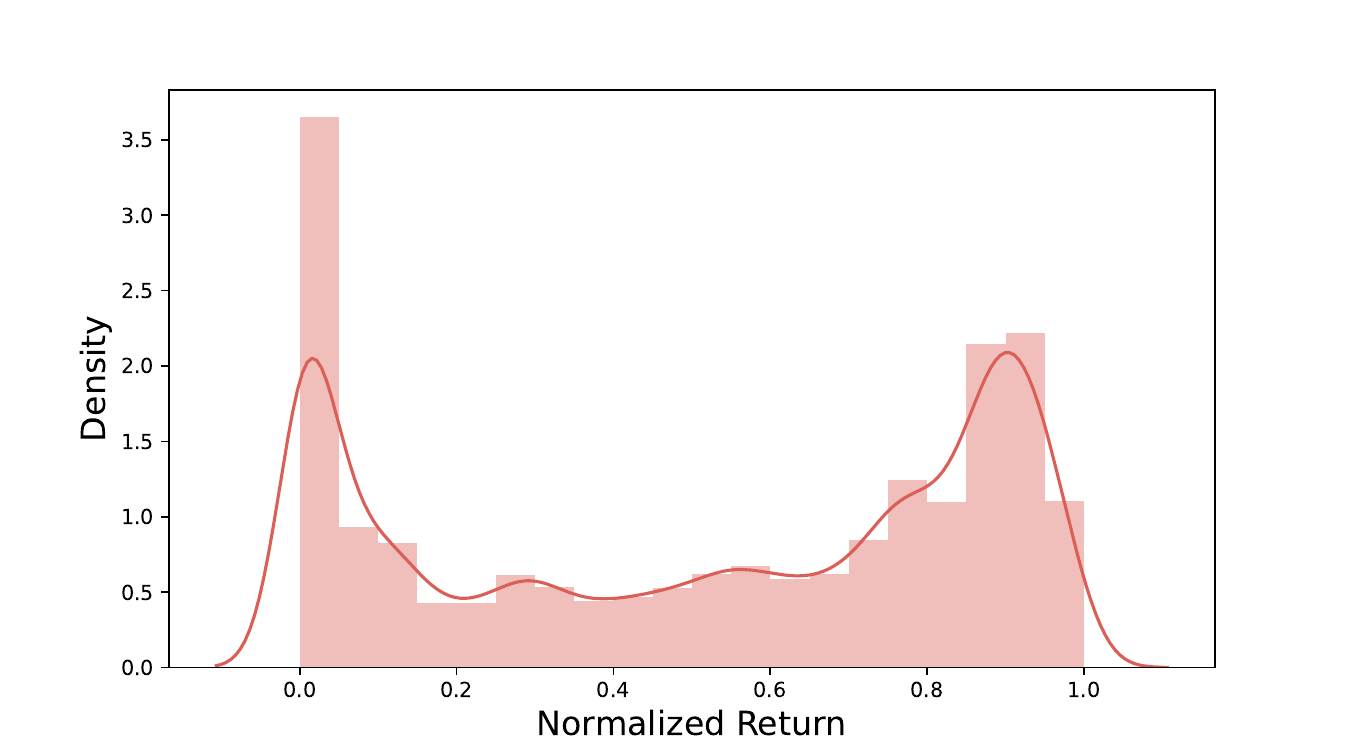}
    }\hspace{-2.3mm}
\subfigure[Sub-optimal dataset]{
	\includegraphics[width=0.49\linewidth]{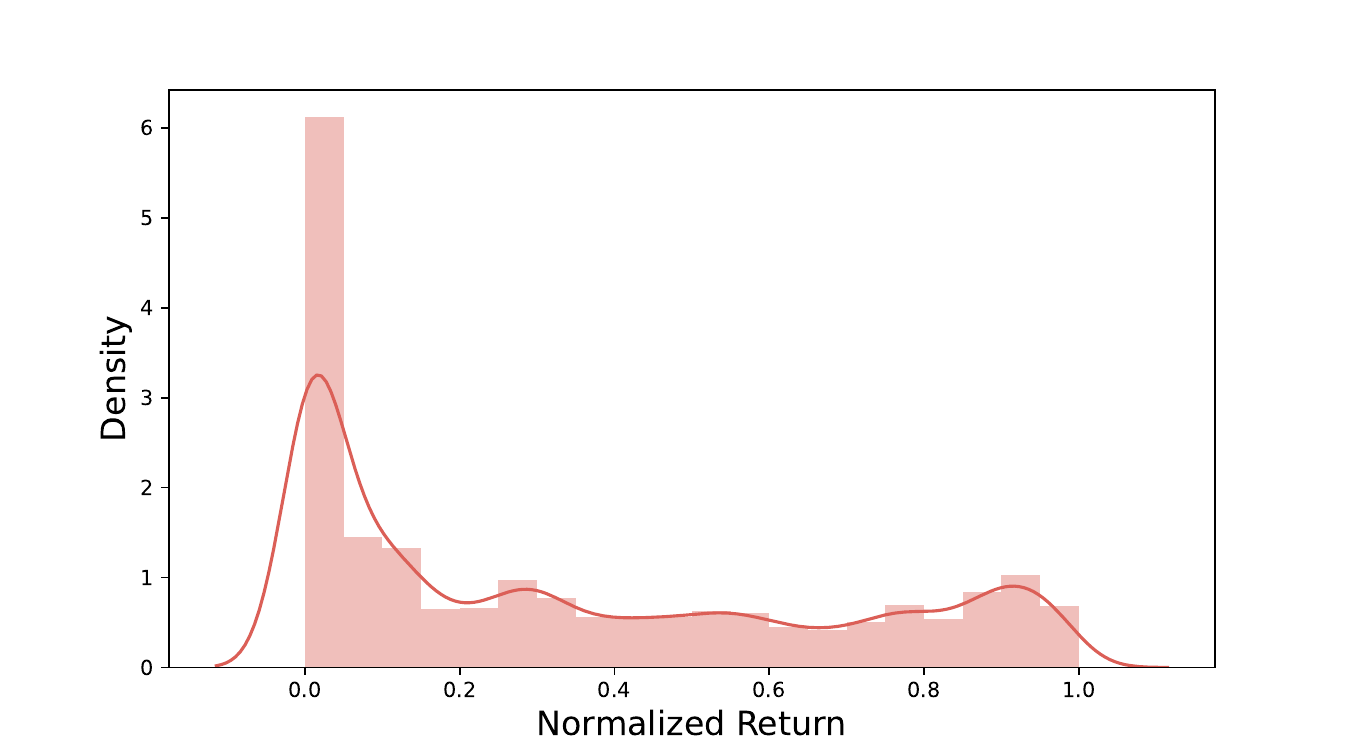}
 }
    \caption{Density visualization of the normalized return in the dataset.}
    \label{fig:dataset}
\end{figure}
\paragraph{Meta-World.} We train Soft Actor-Critic (SAC) \cite{sac} policy in isolation for each task from scratch until convergence. Then we collect 1M transitions from the SAC replay buffer for each task, consisting of recording samples in the replay buffer observed during training until the policy reaches the convergence of performance. For this benchmark, we have two different dataset compositions:
 \begin{itemize}[leftmargin=*]
     \item \textbf{Near-optimal} dataset consisting of the experience (100M
transitions) from random to expert (convergence) in SAC-Replay.
     \item \textbf{Sub-optimal} dataset consisting of the initial 50\% of the trajectories (50M transitions) of the near-optimal dataset for each task, where the proportion of expert data decreases a lot.
 \end{itemize}
To visualize the optimality of each dataset clearly, we plot the univariate distribution of return in each kind of dataset in Fig.~\ref{fig:dataset}.  Our dataset is available at \url{https://bit.ly/3MWf40w}.
\paragraph{Maze2D} The offline dataset is collected by selecting random goal locations and using a planner to generate sequences of waypoints by following a PD controller. We borrow the code from \url{https://github.com/Farama-Foundation/D4RL} \cite{d4rl} to generate datasets for 8 training maps. We collect 35k episodes in total.
\section{Differences Between PromptDT and \textsc{MTDiff-p}}
The remarkable superiority of \textsc{MTDiff-p} over PromptDT emerges from our elegant incorporation of transformer architecture and trajectory prompt within the diffusion model framework, effectively modeling the multi-task trajectory distribution. PromptDT is built on Decision Transformer and it is trained in an autoregressive manner, which is limited to predicting actions step by step. However, \textsc{MTDiff-p} leverages the potency of sequence modeling, empowering it to perform trajectory generation adeptly. \textsc{MTDiff-p} has demonstrated SOTA performance in both multi-task decision-making and data synthesis empirical experiments, while PromptDT fails to contribute to data synthesis. Technically, \textsc{MTDiff-p} extends Decision Diffuser \cite{decisiondiffuser} into the multi-task scenario, utilizing classifier-free guidance for generative planning to yield high expected returns. To further verify our claim, we train our model on the publicly available PromptDT datasets \cite{promptdt}, i.e., Cheetah-vel and Ant-dir. These chosen environments have been judiciously selected due to their inherent diversity of tasks, serving as a robust test to validate the capability of multi-task learning. We report the scores (mean and std for 3 seeds) in Table \ref{table:mt_pro}.
\begin{table}[!ht]
    \centering
    \caption{Average scores obtained by \textsc{MTDiff-p} and PromptDT across 3 seeds. We observed that \textsc{MTDiff-p} outperforms PromptDT largely, demonstrating its high efficacy and potency.}
    \vspace{1em}
    \label{table:mt_pro}
    \begin{tabular}{c|c|c}
    \hline
        \textbf{Methods} & \textbf{Cheetah-vel} & \textbf{Ant-dir} \\ \hline
        \textsc{MTDiff-p} & $\bm{-29.09\pm 0.31}$ & $\bm{602.17\pm 1.68}$ \\ \hline
        PromptDT & $\bm{-34.43\pm2.33}$ & $\bm{409.81\pm9.69}$ \\ \hline
    \end{tabular}
\end{table}
\section{Single-Task Performance}
We train one \textsc{MTDiff-p} model on MT50-rand and evaluate the performance for each task for 50 episodes. We report the average evaluated return in Table ~\ref{table:raw_rate}.
 \begin{table}[!h]
  \begin{center}
    \caption{Evaluated return of \textsc{MTDiff-p} for each task in MT50-rand. We report the mean and standard deviation for 50 episodes for each task.}
    \resizebox{\linewidth}{!}{
    \label{table:raw_rate}
    \begin{tabular}{c|cc} 
    \hline
   \textbf{Tasks}& \textbf{Return on near-optimal dataset} &\textbf{Return on sub-optimal dataset}\\
    \hline
      basketball-v2 & $2735.7\pm1927.7$& $2762.7\pm1928.8$\\
      bin-picking-v2& $733.7\pm1211.8$&$59.6\pm32.3$\\
      button-press-topdown-v2&$1491.6\pm390.9$&$1395.8\pm332.6$\\
      button-press-v2&$2419.2\pm413.7$&$2730.4\pm514.3$\\
      button-press-wall-v2&$3474.7\pm887.6$&$2613.1\pm906.8$\\
      coffee-button-v2&$3157.9\pm1274.3$&$1649.4\pm869.5$\\
      coffee-pull-v2&$437.5\pm597.3$&$98.8\pm115.4$\\
      coffee-push-v2&$463.7\pm729.1$&$71.3\pm55.7$\\
      dial-turn-v2&$2848.1\pm996.3$&$2244.5\pm881.5$\\
      disassemble-v2&$369.9\pm265.4$&$372.3\pm623.2$\\
      door-close-v2&$4325.2\pm377.6$&$4270.4\pm460.1$\\
      door-lock-v2&$3215.1\pm857.5$&$3082.4\pm1041.4$\\
      door-open-v2&$3458.1\pm960.5$&$2457.0\pm833.0$\\
      door-unlock-v2&$2082.6\pm1568.4$&$3078.8\pm1158.1$\\
      hand-insert-v2&$927.9\pm1527.6$&$288.8\pm932.1$\\
      drawer-close-v2&$4824.5\pm8.8$&$4825.3\pm22.3$\\
      drawer-open-v2&$3530.4\pm1512.6$&$2218.8\pm529.3$\\
      faucet-open-v2&$3877.4\pm598.1$&$4245.6\pm575.2$\\
      faucet-close-v2&$4167.1\pm418.8$&$4624.1\pm107.5$\\
      handle-press-side-v2&$3423.3\pm1075.6$&$3995.8\pm1130.0$\\
      handle-press-v2&$3397.9\pm1752.1$&$3125.4\pm1724.7$\\
      handle-pull-side-v2&$3141.5\pm1775.4$&$1474.2\pm1392.9$\\
      handle-pull-v2&$2845.7\pm1970.6$&$2856.8\pm1699.8$\\
      lever-pull-v2&$3383.5\pm1299.2$&$3125.6\pm1433.1$\\
      peg-insert-side-v2&$915.1\pm1624.8$&$497.8\pm1015.3$\\
      pick-place-wall-v2&$649.8\pm1288.6$&$288.2\pm988.1$\\
      pick-out-of-hole-v2&$1792.8\pm1959.9$&$700.5\pm1190.3$\\
      reach-v2&$4144.2\pm645.4$&$966.4\pm842.9$\\
      push-back-v2&$97.2\pm611.3$&$566.3\pm1042.9$\\
      push-v2&$142.0\pm335.7$&$64.6\pm98.2$\\
      pick-place-v2&$166.0\pm759.2$&$7.3\pm4.8$\\
      plate-slide-v2&$4096.4\pm1202.5$&$4306.3\pm495.8$\\
      plate-slide-side-v2&$2910.3\pm616.1$&$2989.0\pm1044.1$\\
      plate-slide-back-v2&$4378.5\pm373.0$&$3963.4\pm927.3$\\
      plate-slide-back-side-v2&$3872.0\pm1151.6$&$4186.5\pm772.6$\\
      soccer-v2&$443.5\pm785.5$&$480.4\pm994.8$\\
      push-wall-v2&$873.9\pm1718.3$&$705.2\pm1535.7$\\
      shelf-place-v2&$204.5\pm714.3$&$366.1\pm919.0$\\
      sweep-into-v2&$1297.9\pm1661.5$&$506.2\pm1368.5$\\
      sweep-v2&$1397.1\pm1922.3$&$599.1\pm1359.2$\\
      window-open-v2&$1453.6\pm1101.4$&$3095.5\pm751.1$\\
      window-close-v2&$2963.9\pm875.3$&$3177.6\pm737.8$\\
      assembly-v2&$2470.7\pm1758.6$&$663.7\pm28.7$\\
      button-press-topdown-wall-v2&$1270.9\pm214.5$&$1199.9\pm203.7$\\
      hammer-v2&$868.7\pm983.4$&$1024.4\pm1024.3$\\
      peg-unplug-side-v2&$1439.1\pm1817.8$&$136.4\pm508.6$\\
      reach-wall-v2&$4249.9\pm536.0$&$4191.4\pm412.9$\\
      stick-push-v2&$1288.5\pm1587.3$&$790.1\pm1097.6$\\
      stick-pull-v2&$601.4\pm1346.3$&$287.1\pm922.8$\\
      box-close-v2&$2683.7\pm1823.4$&$2273.3\pm1707.3$\\
      \hline
    \end{tabular}
    }
  \end{center}
\end{table}

\end{document}